\documentclass[lettersize,journal]{IEEEtran}
\usepackage{amsmath,amsfonts}
\usepackage{algorithmic}
\usepackage{algorithm}
\usepackage{array}
\usepackage[caption=false,font=normalsize,labelfont=sf,textfont=sf]{subfig}
\usepackage{textcomp}
\usepackage{stfloats}
\usepackage{url}
\usepackage{verbatim}
\usepackage{graphicx}
\usepackage{cite}
\usepackage{color}
\newcommand{\red}[1]{\textcolor{red}{#1}}
\usepackage{multirow}
\usepackage{makecell}

\begin{document}

\title{Enhanced Long-Tailed Recognition with Contrastive CutMix Augmentation}

\author{Haolin Pan*, Yong Guo*, Mianjie Yu* and Jian Chen$^\dag$

\thanks{Haolin Pan is with the South China University of Technology (SCUT), Guangzhou, 510006, China (email: mr.haolinpan@qq.com).}
\thanks{Yong Guo is with the Max Planck Institute for Informatics (MPI-INF), Saarbrücken, 66123, Germany (email: guoyongcs@gmail.com).}
\thanks{MianJie Yu is with the University of Macau (UM), Macau SAR, 519000, China (email: mianjieyu@gmail.com).}
\thanks{Jian Chen is with the South China University of Technology (SCUT), Guangzhou, 510006, China (email: ellachen@scut.edu.cn).}
\thanks{\emph{* Authors contributed equally. $^\dag$  Corresponding author.}}
}

\markboth{IEEE TRANSACTIONS ON IMAGE PROCESSING,~Manuscript}%
{Shell \MakeLowercase{\textit{et al.}}: A Sample Article Using IEEEtran.cls for IEEE Journals}

\maketitle

\begin{abstract}
Real-world data often follows a long-tailed distribution, where a few head classes occupy most of the data and a large number of tail classes only contain very limited samples. In practice, deep models often show poor generalization performance on tail classes due to the imbalanced distribution. To tackle this, data augmentation has become an effective way by synthesizing new samples for tail classes. Among them, one popular way is to use CutMix that explicitly mixups the images of tail classes and the others, while constructing the labels according to the ratio of areas cropped from two images.
However, the area-based labels entirely ignore the inherent semantic information of the augmented samples, often leading to misleading training signals.
To address this issue, we propose a Contrastive CutMix (ConCutMix) that constructs augmented samples with semantically consistent labels to boost the performance of long-tailed recognition.
Specifically, we compute the similarities between samples in the semantic space learned by contrastive learning, and use them to rectify the area-based labels.
Experiments show that our ConCutMix significantly improves the accuracy on tail classes as well as the overall performance. For example, based on ResNeXt-50, we improve the overall accuracy on ImageNet-LT by 3.0\% thanks to the significant improvement of 3.3\% on tail classes. We highlight that the improvement also generalizes well to other benchmarks and models.
Our code and pretrained models are available at https://github.com/PanHaulin/ConCutMix.
\end{abstract}

\begin{IEEEkeywords}
long-tailed recognition, data augmentation, contrastive learning.
\end{IEEEkeywords}

\section{Introduction}
\begin{figure}[t]
    \centering
\centerline{\includegraphics[width=1\columnwidth]{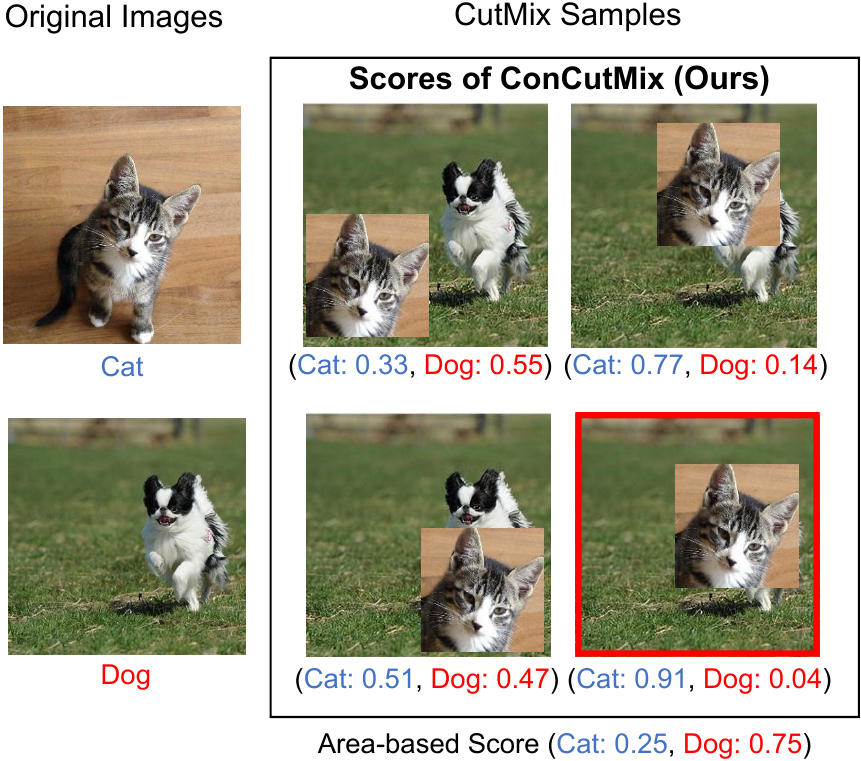}}
    \caption{Difference between the area-based labels from CutMix and the scores of ConCutMix. These images are synthesized with the same area ratio but have different semantics. The image with a \textcolor{red}{red} box shows that the area-based label may be entirely wrong in terms of semantics. In contrast, the scores for ConCutMix are intuitively more consistent with semantics.}
    \label{fig:area_sim}

\end{figure}

Unlike the standard way of training models on balanced data, real-world datasets often follow long-tailed distributions, with a few classes dominating the majority of the dataset (ie., head classes), while other classes have a small number of samples (tail classes).
This imbalance between head and tail classes presents difficulties in training deep models that can make unbiased decisions across classes. 
Data augmentation is a widely adopted technique to address the issue of imbalanced data 
by synthesizing new samples for tail classes~\cite{chou2020remix,baik2022dbn,xu2021towards,tiong2021improving}.
Compared to oversampling~\cite{peng2020large,hu2020learning}, data augmentation alleviates the scarcity of tail classes in a cost-effective manner, thereby enhancing the model's generalization performance.
Recently, CutMix~\cite{yun2019CutMix} has attracted more attention as it enables increasing the diversity of samples in tail classes by explicitly mixing the images of tail classes with other classes~\cite{jeong2022supervised,park2022majority}.
CMO~\cite{park2022majority} simply transfers the rich context of head classes to tail classes by CutMix, yielding the state-of-the-art performance on long-tailed recognition tasks. 

\begin{figure*}[t]
    \centering
\centerline{\includegraphics[width=2\columnwidth]{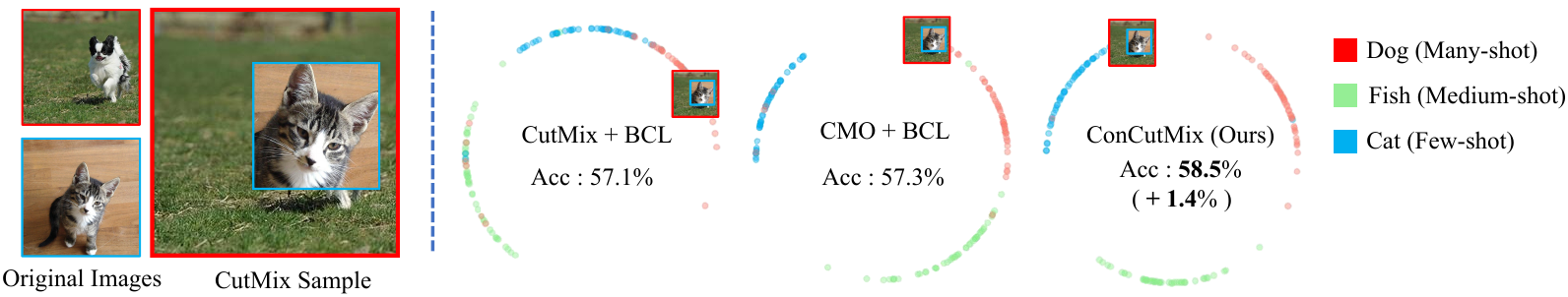}}
    \caption{Visualization of the feature distribution on ImageNet, with each color representing a specific class. We compare ConCutMix with two popular augmentation methods, i.e., CutMix~\cite{yun2019CutMix} and CMO~\cite{park2022majority}, all of which are implemented on BCL~\cite{zhu2022balanced}. Clearly, our ConCutMix can effectively separate the classes with a significant difference in the number of samples. Besides, ConCutMix distributes the considered synthetic sample to a semantically appropriate position, {whereas other methods distribute it near classes that are clearly semantically inconsistent}. 
    }
    \label{fig:feat}
\end{figure*}

\begin{figure}[t]
    \centering
\centerline{\includegraphics[width=1\columnwidth]{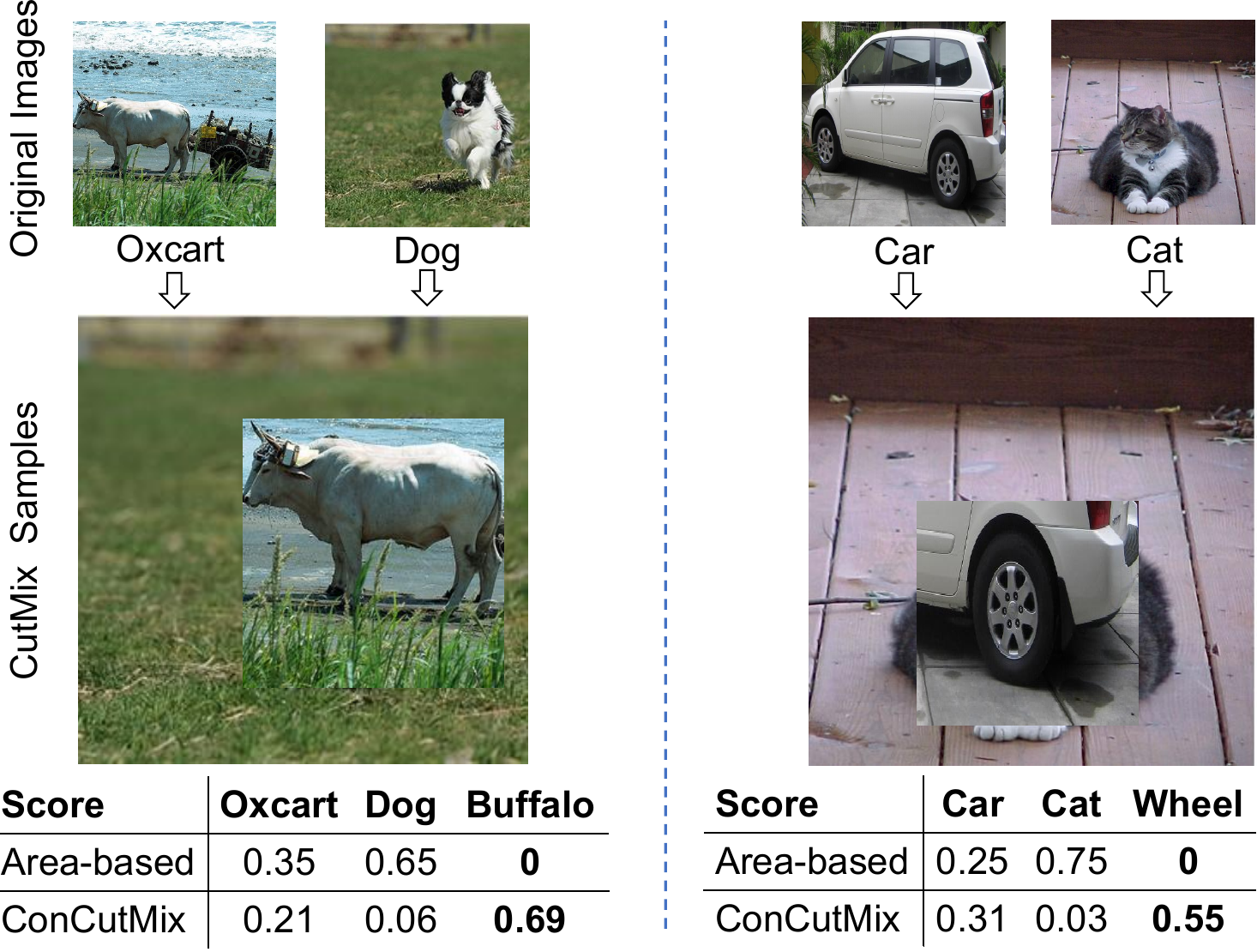}}
    \caption{
    Difference between the area-based labels and the scores of ConCutMix, in treating novel classes that are not used for CutMix. We show two synthetic samples may semantically belong to a novel class other than the considered classes in CutMix, while ConCutMix is able to capture the semantically consistent information for each synthetic sample.}
    \label{fig:ox}
\end{figure}

\IEEEpubidadjcol
However, we discover that these methods still show limited performance since they neglect the semantic information of the synthetic samples. To be specific, they simply construct area-based labels based on the ratio of areas cropped from two considered images.
As a result, the area-based labels may entirely mismatch the real semantic information of the augmented examples in at least two ways. First, as shown in Fig.~\ref{fig:area_sim}, if we mixup a cat image and a dog image with the area ratio of 1:3, the label would always be (cat:0.25, dog:0.75) no matter how we combine these two images. This label could be entirely wrong if we use the cat patch to cover the body of the dog (highlighted by the \textcolor{red}{red} box). In this case, this example should be a cat instead of a dog, indicating that the area-based label becomes very misleading.
Second, the augmented samples mixed from images of two classes could possibly belong to a novel class other than the considered two classes, as illustrated by the examples in Fig.~\ref{fig:ox}. To be specific, on ImageNet, when mixing images of ``Oxcart'' and ``Dog'', we could obtain an image that belongs to a novel class ``Buffalo''. 
This phenomenon also occurs in another example where mixing images of "Car" and "Cat" results in an image of "Wheel."
Based on the aforementioned observations, we highlight that the area-based labels frequently exhibit inconsistencies with the real semantics of images and thereby could mislead the training process.

To address this, we propose a Contrastive CutMix (ConCutMix) to construct synthetic samples with semantically consistent labels.
Specifically, we follow a recent work, BCL~\cite{zhu2022balanced}, to learn a semantically discriminative feature space using supervised contrastive learning. Then, we measure the similarities between synthetic samples and the center of each class in the feature space to construct semantically consistent labels. 
After that, we combine the semantically consistent label with the standard area-based label to alleviate the issue of misleading signals.
As mentioned above, since the synthetic image could belong to a novel class beyond the classes of two input images, we propose to select the TopK (with $K{>}2$) classes when computing similarities in the learned feature space and use them to construct our semantically consistent label.
To verify the proposed ConCutMix, we visualize the distribution of examples in the learned feature space in Fig.~\ref{fig:feat}. For simplicity, we select three representative classes for visualization, including one majority class, one medium class, and one minority class. When taking the extreme case that the area-based label entirely fails in Fig.~\ref{fig:area_sim}, both CutMix~\cite{yun2019CutMix} and CMO~\cite{park2022majority} still recognize it as ``Dog'' but our ConCutMix successfully put it towards the cluster of ``Cat'', yielding semantically consistent labels.

We highlight that, ConCutMix benefits the tail classes by enhancing the \textbf{\emph{quality}} and \textbf{\emph{quantity}} of data (labels) associated with them.
Firstly, ConCutMix improves the label quality of samples in tail classes by alleviating semantic inconsistencies. 
Suppose ``Cat'' is a tail class, for the example with the \textcolor{red}{red} box in Fig.~\ref{fig:area_sim}, the area-based label would assign a high weight for the background class ``Dog'', yielding a misleading label. By contrast, the semantically consistent label effectively preserves the information of ``Cat'', thereby providing a label of superior quality.
Secondly, ConCutMix is able to produce additional high-quality data for tail classes. Specifically, suppose ``Buffalo'' is a tail class in Fig.~3 (left), we can generate new images of a tail class from two head classes (e.g., Oxcart and Dog) via CutMix. However, CutMix only incorporates information from either head class, while ConCutMix is capable of providing the correct label of ``Buffalo'', yielding a new training sample.

\begin{figure*}[h]
    \centering
\centerline{\includegraphics[width=2\columnwidth]{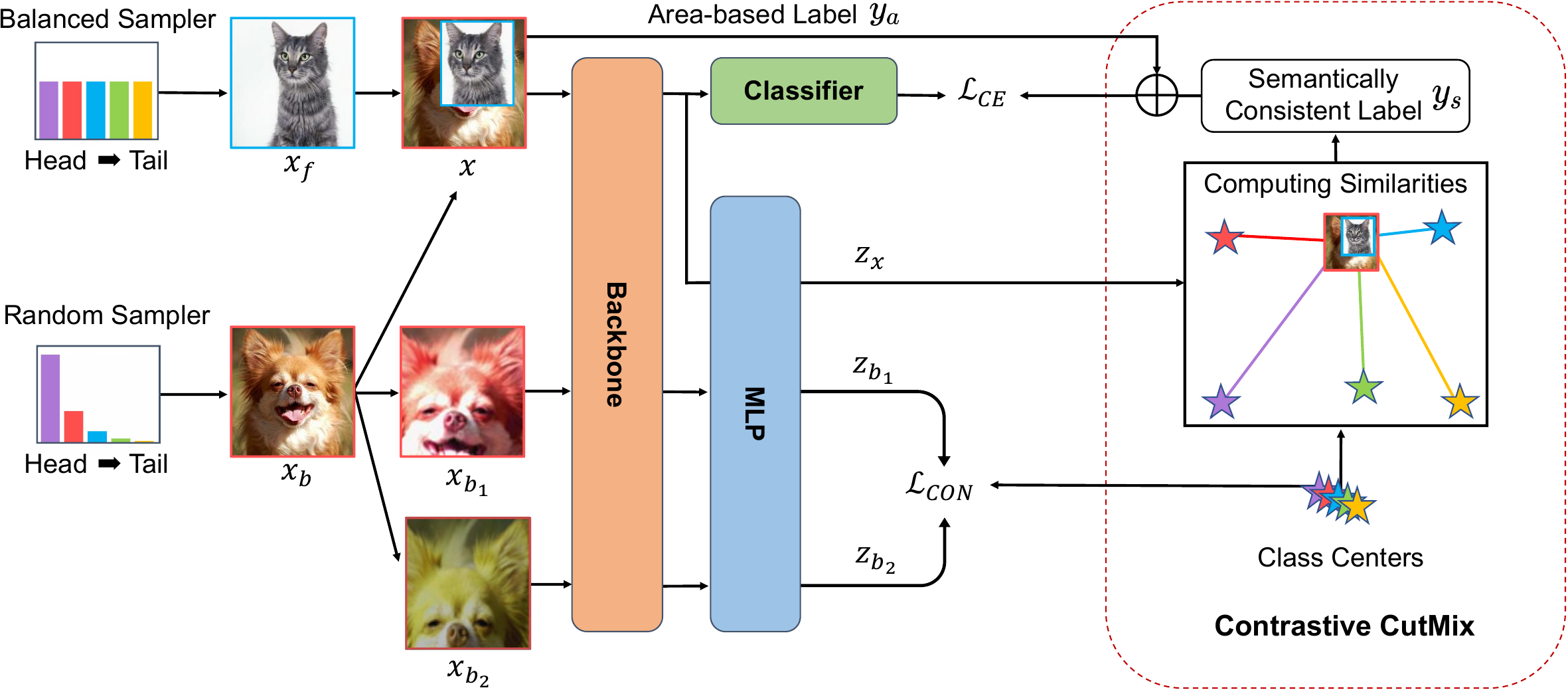}}
    \caption{Overall pipeline for long-tailed recognition with ConCutMix. We first perform CutMix with images sampled from a balanced sampler and a random sampler to synthesize samples. Balanced sampler expands tail classes by increasing the probability of being foreground images from them. Following BCL~\cite{zhu2022balanced}, we leverage contrastive learning to establish a semantic space, where ConCutMix can construct semantically consistent labels based on the similarities with learned class centers. Finally, ConCutMix simply incorporates semantically consistent labels into training to rectify the area-based labels generated by CutMix.} 
    \label{fig:frame}
\end{figure*}

Our main contributions can be summarized as follows:
\begin{itemize}

    \item We propose a novel augmentation method for long-tailed recognition, dubbed Contrastive CutMix, which provides semantically consistent labels for the samples generated by CutMix. To rectify the area-based labels, we explicitly compute the similarity between the synthetic sample and different class centers/prototypes in the semantic space learned by contrastive learning.

    \item Interestingly, we notice that mixing images from two classes may not necessarily obtain an augmented sample belonging to them. 
    To address this issue, we consider TopK-similar classes to construct the resultant semantically consistent label for each sample.

    \item In experiments, the proposed ConCutMix consistently improves the performance by a large margin across diverse benchmarks. Specifically, with the imbalance factor of 100, we obtain an improvement of 1.76\% and 1.15\% on CIFAR-10-LT and CIFAR-100-LT, respectively. We also improve the accuracy on ImageNet-LT by 3.0\% and iNaturalist 2018 by 1.0\%.

\end{itemize}

\section{Related Work}
\noindent\textbf{Long-Tailed Recognition.}
Real-world data typically follows a long-tailed or imbalanced distribution, which biases the learning toward head classes, and degrades the performance on tail classes  \cite{yang2022multi,zhang2021deep,colearning}. Over-sampling \cite{buda2018systematic,shen2016relay} is one of the most common methods to alleviate data scarcity, it emphasizes the tail classes and increases the instance number of the tail classes to reduce the imbalance.
{Re-weighting methods adjust the loss weights for different classes \cite{byrd2019effect,cao2019learning,cui2019class} or adjust the logits~\cite{menon2020long,InverseAlex} during
training to improve the tail class.}
Featured-based methods \cite{abcnorm,pcl} often introduce additional losses to improve the ability of the model to recognize tail classes.
Data augmentation \cite{park2022majority,chou2020remix,li2021metasaug} is another way of data processing to solve
the problem of long-tailed distribution.
Therefore the tail classes can be compensated by data augmentation methods. However, these methods do not adjust the labels but we conduct the semantically consistent label to improve the performance of the tail class with oversampling.

\noindent\textbf{Data Augmentation.}
Recently, data augmentation has performed satisfactorily in the computer vision, such as action recognition~\cite{CutMixLiu,SampleMeng}.
Cutout \cite{devries2017improved} removes random regions whereas Mixup \cite{zhang2017mixup} linearly interpolates two images in a training dataset.
A more general and powerful data augmentation technique, known as CutMix~\cite{yun2019CutMix}, involves explicitly combining Cutout and Mixup. Specifically, CutMix fills the removed regions with patches from another training image and mixes their corresponding labels. 
{Recent works have focused on automatically searching for strategies for image transformations~\cite{AutoTang,diffaug}. 
But most advanced long-tailed methods still use Mixup or CutMix, as they effectively augment scarce samples and benefit the tail classes.}
Zhou et al.  \cite{zhou2020bbn} use the mixup to improve calibration for long-tailed recognition, and MiSLAS  \cite{zhong2021improving} uses mixup in the first training stage to improve the tail class by re-sampling. 
Remix \cite{chou2020remix} assigned a label in favor of the minority classes when mixing two samples. However, these methods apply augmentation neglecting the semantic information of synthetic samples, just simply constructing corresponding area-based labels, which will cause the limitation of the tail class with oversampling. 
Unlike these methods, we sample images from different distributions and provide semantically consistent labels for the samples to break the limitation of class composition from CutMix.
 
\noindent\textbf{Contrastive Learning.} 
Recent studies have demonstrated that contrastive learning helps achieve strong performance in representation learning \cite{deng2022boosting,cacl,CLAST}.
\cite{hyvarinen2005estimation} proposes and refines Noise
Contrastive Estimation (NCE). \cite{oord2018representation} proposes InfoNCE to help the model capture valuable information for future prediction. 
Supervised contrastive learning (SupCon)  \cite{khosla2020supervised} extends contrastive learning to the fully-supervised setting.
Recently, researchers have applied contrastive learning to imbalanced and long-tailed classification and demonstrated improved performance. 
{SSP \cite{yang2020rethinking} achieves balanced feature spaces by incorporating contrastive learning into both semi-supervised and self-supervised methods.}
Hybrid-SC \cite{wang2021contrastive} and Hybrid-PSC \cite{wang2021contrastive} design a two-branch
network, using a contrastive learning branch and a classifier branch for eliminating the bias. 
KCL \cite{kang2020exploring} adopts the two-stage learning paradigm and provides the same number of positives for all classes in every batch.
TSC \cite{li2022targeted} improves the KCL by urging the features of classes closer to the target features on the vertices of a regular simplex. PaCo \cite{cui2021parametric} introduces a set of class-wise learnable centers. BCL \cite{zhu2022balanced} also uses the learnable class center and weighting method to improve the tail classes. However, these methods seek to improve training methods for long-tailed recognition, while ConCutMix focuses on another perspective, i.e., improving the quality and quantity of data in tail classes by augmentation.

\begin{figure*}[ht]
    \centering
\centerline{\includegraphics[width=2\columnwidth]{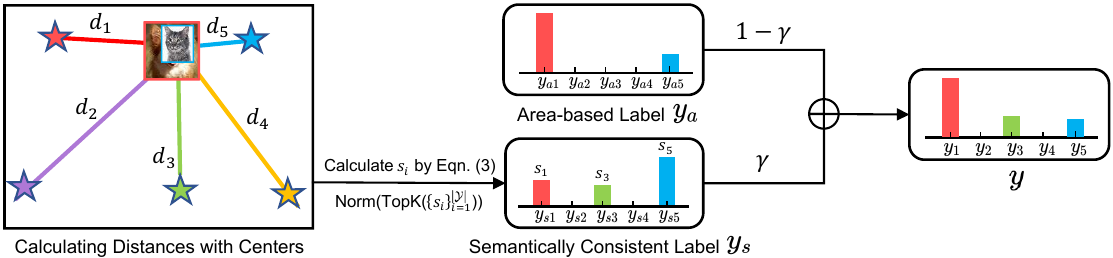}}
\vspace{-5 pt}
    \caption{
    Details of rectifying area-based labels with semantically consistent labels. We first calculate similarities between a synthetic sample and all learned class centers. ConCutMix constructs semantically consistent labels based on the similarities with the TopK-similar classes. 
    To alleviate the misleading from noisy semantic information due to sample scarcity of tailed classes, ConCutMix combines normalized semantically consistent label with area-based label under the control of confidence function $\gamma(\cdot)$. }
    \label{fig:detail}
    \vspace{-5 pt}
\end{figure*}

\section{Proposed Method}
\subsection{{Preliminary and Notation}}

\noindent
\textbf{CutMix and Label Construction.}
Let  $x \in \mathbb{R}^{W \times H \times C}$, $y  \in \mathcal{Y} $ denote an image and its label. 
CutMix randomly samples two images as the foreground $(x_f, y_f)$ and background $(x_b, y_b)$, respectively. These two images are linearly combined to form a new sample $({x}, {y})$:
\begin{equation}\label{eq:CutMix}
\begin{split}
&{x} =\mathbf{M} \odot x_{f}+(\mathbf{1}-\mathbf{M}) \odot x_{b}, \\
&{y} =\lambda \cdot y_f+(1-\lambda) \cdot y_b,   
\end{split}
\end{equation}
where $\mathbf{M} \in\{0,1\}^{W \times H}$ denotes a binary mask indicating
the position to drop and fill from two images, \textbf{1} is a binary mask filled with ones and $\odot$ is element-wise multiplication. $\lambda \sim \mathcal{U}(0,1)$ is sampled from a uniform distribution, indicating the area ratio of the foreground to the entire image. 

\begin{figure*}[h]
\centering
    \includegraphics[width=2\columnwidth]{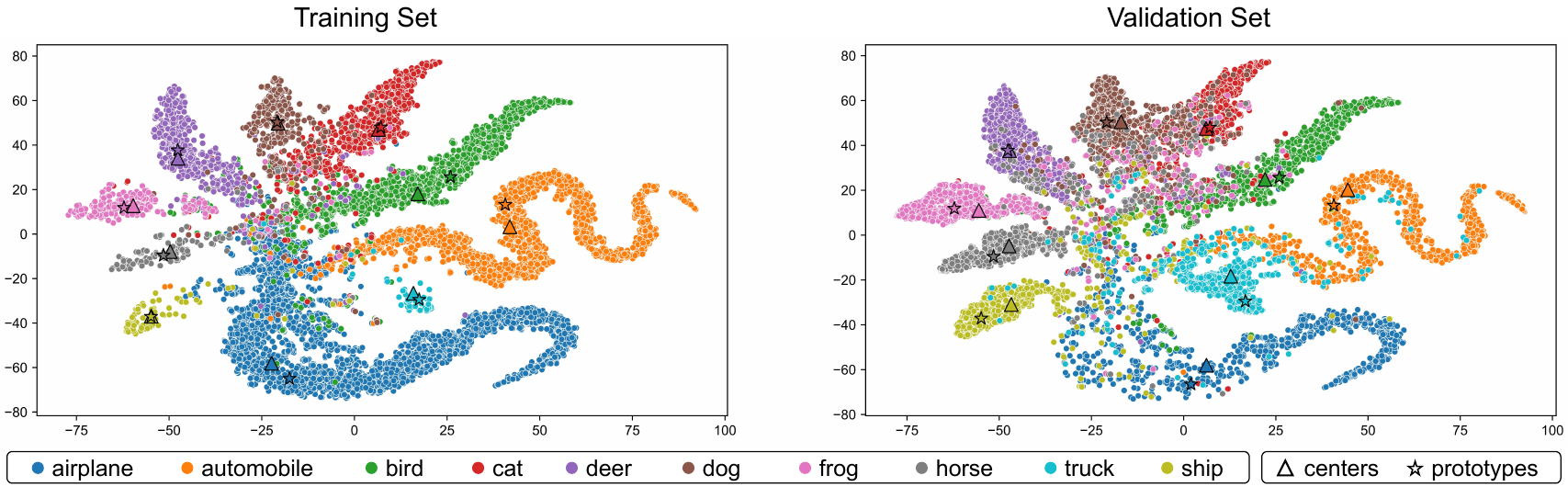}
    \caption{{Visualization of the feature space learned by ConCutMix on CIFAR-10-LT with an imbalance factor of 100, where each color represents a specific class. Prototypes learned from contrastive learning are denoted as stars. The class centers calculated by means are represented by triangles. Clearly, in this well-learned semantic space, the learned prototypes are sufficiently representative to serve as class centers for both the training and validation sets.}}
    \label{fig:tsne}
\end{figure*}

\noindent
\textbf{Learning Semantic Feature Space.}
In this paper, we seek to capture semantic information of examples to provide reliable labels. One possible way is to learn a semantic feature space and measure the distance/similarity between examples within it. To this end, we mainly follow~\cite{zhu2022balanced} and use the prototypes based contrastive learning methods to learn the feature space.
{
Typically, contrastive learning projects samples to representations $z$ using a nonlinear multiple layer perception (MLP) that follows the backbone. The representations $z$ are subsequently L2-normalized to yield $\bar{z}$.
$A_l$ includes a set of $\bar{z}$ and a corresponding prototype $c_l$ belonging to the class $l$. $P_i$ denotes the positive set of $x_i$ obtained by removing $\bar{z}_i$ from $A_{y_i}$.
Consequently, the supervised contrastive loss with learnable centers~\cite{zhu2022balanced} can be formulated as:
\begin{equation}\label{con}
\mathcal{L}_{\mathrm{CON}}(x_i) =-  \frac{1}{|P_i|}{
\sum\limits_{\bar{z}_p\in P
_i}\log\dfrac{\exp(\bar{z}_i \cdot \bar{z}_p/\tau)}{\sum\limits_{l\in\mathcal{Y}}
\frac{1}{|A_l|}
 \sum\limits_{\bar{z}_j\in A_l}\exp(\bar{z}_i\cdot \bar{z}_j/\tau)}
},
\end{equation}
where $|\cdot|$ is the size of a set and $\tau$ is a a scalar parameter named temperature.
}
As shown in \eqref{con}, prototypes are paired with the samples of the same class as positive pairs but paired with others as negative pairs. As a result, each sample will be pulled toward corresponding prototypes while being pushed away from others.
Interestingly, \cite{graf2021dissecting} demonstrated that supervised contrastive learning facilitates samples of each class collapsed to the vertices of a regular simplex in the feature space. From these points of view, if the supervised contrastive loss converges well, the learned prototypes can be regarded as representative class centers. {
To illustrate this, we visualize the features of CIFAR-10-LT on both the training and validation sets in Fig.~\ref{fig:tsne}, marking the prototypes with stars. And the class centers, which are calculated using means, are marked with triangles. It can be observed that the prototype and class center belonging to the same class are closely pulled together by contrastive learning. Although the distance between them increases slightly on the validation set compared to the training set, it is still sufficient to demonstrate that in a semantically separable feature space, prototypes are representative enough to serve as learned class centers.}

\subsection{Contrastive CutMix}
In this paper, we observed that CutMix often produces misleading area-based labels that are inconsistent with the real semantics of examples.
To address this issue, we propose an augmentation method, dubbed Contrastive CutMix~(ConCutMix), which seeks to construct semantically consistent labels to rectify the area-based labels.
In this part, we first elaborate on the details of semantically consistent label construction. 
Then, we further explain how to rectify the area-based labels with semantically consistent labels. 
The overview of the proposed ConCutMix is shown in Fig.~\ref{fig:frame}.

\noindent\textbf{Constructing Semantically Consistent Labels by Contrastive Learning.}
{Class centers for calculating the semantics of synthetic samples can be obtained in several ways. For instance, class centers can be calculated using means at each iteration or periodically updated using an exponential moving average [41]. However, we advocate for the use of prototype-based contrastive learning~\cite{cui2021parametric,zhu2022balanced}, as it creates a discriminative semantic feature space where class prototypes can act as learnable class centers without additional computational costs. In this learned semantic space, we can measure the distance or similarity between samples and different classes.}
To achieve this goal, we first compute the Euclidean distance between an synthetic sample and each class center in the feature space.
Then, we follow~\cite{yang2011spectral} and use the reciprocal of Euclidean distance to calculate the similarity, i.e., a larger distance means a lower similarity.
Formally, the similarity between synthetic samples ${x}$ and the $i$-th class center $c_i$ is:
\begin{equation}
    s({x},c_i)=\frac{1}{d({x},c_i)},\quad d({x},c_i)=||z_{{x}}-c_i||_2.
\end{equation}
Besides Euclidean distance, the similarities can also be measured by cosine similarity. We provide a further comparison between these two measurements in Section~\ref{discussion}.
Naturally, we can construct semantically consistent labels by only considering the similarities between the two classes used for mixing images, while setting the similarities for other classes to 0.  
However, we observe that the synthetic image could belong to a novel class beyond the considered two classes. As illustrated in Fig.~\ref{fig:ox}, when combining images of ``Oxcart'' and ``Dog'', a new category called ``Buffalo'' could be created. Similarly, the merging of images of ``Car'' and ``Cat'' results in an image of a ``Wheel''. 
To handle this, we propose to construct the semantically consistent labels by considering TopK-similar classes.
Formally, our semantically consistent label can be computed by:
\begin{equation}
    y_s =\ell\big(\mathrm{TopK}(\{s({x},c_i)\}_{i=1}^{\mathcal{|Y|}})\big),
    \label{eq:topk}
\end{equation}
where $\ell(\cdot)$ is a normalization to ensure the sum of weights of classes is equal to one.
We empirically set the number of associated classes as $\mathrm{K}=0.3\mathcal{|Y|}$, but the maximum does not exceed $30$, which will be further discussed in Section~\ref{discussion}. 

\noindent\textbf{Rectifying Area-based Labels with the Proposed Semantically Consistent Labels.}
A naive way to rectify the noisy training signals is completely replacing area-based labels with semantically consistent labels.
However, it shows limited performance as semantically consistent labels may hamper the training process in the early training stage.
To be specific, the learned class centers are not good enough when a discriminative feature space has not been fully established in the early stage, resulting in noisy semantic feature space. To alleviate this issue, as shown in Fig.~\ref{fig:detail}, we seek to stable the semantically consistent labels $y_s$ with the area-based labels $y_a$, using a weighted sum manner:
\begin{equation}
    y = \big(1-\gamma({x})\big) \cdot y_a + \gamma({x}) \cdot y_s, 
\end{equation}
where $\gamma(\cdot)$ is a confidence function to balance between these two labels. 
Intuitively, the confidence of $y_s$ should be determined by the reliability of its semantic information in the learned semantic feature space. 
For long-tail recognition, we highlight that the reliability should vary a lot among different classes.
To be specific, semantic information from the head classes is considered reliable because they are learned from sufficient data. In contrast, the semantic information captured from the tail classes is often biased due to data scarcity.

Inspired by this, we estimate the reliability of semantic information according to how much data are available for the two basic classes that the synthetic example is mixed from. For example, as shown in \eqref{eq:CutMix}, the confidence of the synthetic sample $x$ should rely on the number of samples in the foreground class $y_f$ and the number of samples in the background class $y_b$, which are denoted by $N_{y_f}$ and $N{y_b}$, respectively. Since the synthetic sample $x$ is built from both $y_f$ and $y_b$, we directly measure its reliability using a weighted sum of $N_{y_f}$ and $N_{y_b}$, with the weight being the area ratio $\lambda$ used by CutMix: 
\begin{equation}
    N_{x} =  \lambda  N_{y_{f}} + (1-\lambda)  N_{y_{b}}.
\end{equation}
Ideally, we should give larger confidence/reliability scores to those samples built from the head classes, while the opposite for the tail classes. To quantify this, we use an activation function $\varphi$ to project the reliability to a numerical score. In this way, the overall confidence can be measured by the ratio of the reliability score of a single sample $\varphi(N_x)$ to the reliability of the whole dataset across all classes $\sum_{i \in \mathcal{Y}} \varphi(N_i)$. Formally, it can be computed by
\begin{equation}
    \gamma(x)= \frac{\omega \cdot \varphi(N_{x})}{\sum\limits_{i\in\mathcal{Y}} \varphi(N_{i})}, \label{gamma}
\end{equation}
where $\omega$ is a scaling factor.
Actually, the activation function can be any form and here we use the logarithmic function since it works best empirically (see ablation in Table~\ref{tab: weighting method}).

\begin{algorithm}[tb]
   \renewcommand{\algorithmicensure}{\textbf{Output:}}
   \caption{Contrastive CutMix:
   First, we measure similarities with learnable centers.
   Second, we construct semantically consistent labels based on similarities.
   Third, we rectify area-based labels with semantically consistent labels.}
   \label{algorithm}
\begin{algorithmic}[1]

\REQUIRE
   Synthetic sample $({x}, {y})$, number of samples belonging to a class $N$, area ratio $\lambda$,
   label space $\mathcal{Y}$,
   class center ${c}$, 
   activation function $\varphi(\cdot)$,
   scaling factor $\omega$, area-based label $y$, normalization function $\ell(\cdot)$.
\STATE Obtain feature of synthetic sample $z_{{x}}$ by projector.
\FOR{$i$ in $\mathcal{Y}$}
\STATE 
Calculate similarities: $ s({x},c_i)=1/||z_{{x}}-c_i||_2$
\ENDFOR
 \STATE Compute semantically consistent label:\\
~~~~~$     y_s =\ell\big(\mathrm{TopK}(\{s({x},c_i)\}_{i=1}^{\mathcal{|Y|}})\big)$
\STATE 
Calculate confidence of semantically consistent label:  \\
$\quad N_{x}= \lambda  N_{y_{f}} + (1-\lambda)  N_{y_{b}}$\\
$\quad \gamma({x})=\omega \cdot \varphi(N_{x})/\sum_{i\in\mathcal{Y}}\varphi(N_i)$
\STATE  Rectify labels with semantically consistent label: \\
~~~~~$    y = \big(1-\gamma({x})\big) \cdot y_a + \gamma(x) \cdot y_s$
\end{algorithmic}
\end{algorithm}

\noindent\textbf{Training Objective.}
Following BCL~\cite{zhu2022balanced}, we use the same training objective that consists of both cross-entropy loss $\mathcal{L_{\rm CE}}$ and contrastive loss $\mathcal{L_{\rm CON}}$:
\begin{equation}
    \mathcal{L} = \alpha \mathcal{L}_{\mathrm{CON}} + \beta \mathcal{L}_{\mathrm{CE}},
\end{equation}
where $\alpha$ and $\beta$ control the weight of individual losses. We highlight that, for fair comparisons, we use exactly the same hyperparameters as BCL in our experiments.

\begin{table*}
\begin{center}
\caption{Comparisons of Top-1 accuracy of different methods implemented with ResNet-32 on CIFAR-100-LT and CIFAR-10-LT for 200 epochs. These results reveal that ConCutMix significantly boosts BCL~\cite{zhu2022balanced} with various imbalance factors and yields better performances on these long-tailed datasets. 
The values in parentheses show the accuracy improvements of ConCutMix over BCL~\cite{zhu2022balanced}.
The best results are marked in bold.}\label{cifar-table}
\resizebox{0.95\linewidth}{!}{
\begin{tabular}{l|ccc|ccc}
\hline
Method &\multicolumn{3}{c}{CIFAR-100-LT}&\multicolumn{3}{c}{CIFAR-10-LT}\\
\hline
Imbalance Factor    & 100 & 50 & 10  & 100 & 50 & 10 \\ \hline
    Focal Loss  \cite{lin2017focal}  & 38.41 & 44.32 & 55.78 & 70.38 & 76.72 & 86.66 \\
   
    CB-Focal  \cite{cui2019class}  & 39.60  & 45.17 & 57.99 & 74.57 & 79.27 & 87.10 \\
     LDAM-DRW  \cite{cao2019learning}  & 42.04 & 46.62 & 58.71 & 77.03 & 81.03 & 88.16 \\
    SSP  \cite{yang2020rethinking}   & 43.43 & 47.11 & 58.91 & 77.83 & 82.13 & 88.53 \\

    BBN  \cite{zhou2020bbn}   & 42.56 & 47.02 & 59.12 & 79.82 & 81.18 & 88.32 \\
    Casual Model
     \cite{tang2020long}  & 44.10  & 50.30  & 59.60  & 80.60  & 83.60  & 88.50 \\

    KCL \cite{kang2020exploring}& 42.80& 46.30& 57.60&
    77.60 &81.70 &88.00\\
    Hybrid-SC  \cite{wang2021contrastive}  & 46.72 & 51.87 & 63.05 & 81.40  & 85.36 & 91.12 \\
    MetaSAug-LDAM
     \cite{li2021metasaug}  & 48.01 & 52.27 & 61.28 & 80.66 & 84.34 & 89.68 \\

    TSC \cite{li2022targeted} &  43.80&  47.40 &  59.00 &  79.70 & 82.90 &88.70\\  
    ResLT  \cite{cui2022reslt}  & 48.21 & 52.71 & 62.01 & 82.40  & 85.17 & 89.70 \\
    Remix \cite{chou2020remix} & 41.94 & 49.50 & 59.36 &75.36 &- & 88.15   \\
     UniMix \cite{xu2021towards} &  45.45&  51.11 & 61.25 & 82.75 & 84.32 &89.66\\
    SMC \cite{jeong2022supervised} & 48.90 & 52.30&62.50& -& - & -  \\
        \hline
    BCL \cite{zhu2022balanced}     & 52.01 & 56.32 & 64.01 & 84.31 & 87.26 & 90.91 \\
     ~ + Mixup \cite{zhang2017mixup}     & 50.69 & 55.46 & 62.45 & 83.72 & 86.85 & 91.00 \\
     ~ + EWB  \cite{hasegawa2023exploring}     & 51.19	&56.49&	63.98	&84.91&	87.31	&90.98 \\
     ~ + WD  \cite{WD}     & 52.39&	56.41	&63.78	&82.91	&86.42&	89.81
 \\ 
      ~ + CR  \cite{CR}     &52.21	&56.41	&53.91	&84.41	&87.12	&90.83

 \\ 
     ~ + RankMix  \cite{RankMix}     & 52.15	&56.81	&63.91	&84.56	&87.41&	91.02
 \\ 
    ~ + CutMix \cite{yun2019CutMix}     & 52.04 & 56.50 & 63.53 & 84.41 & 87.37 & 91.05 \\
     ~ + CMO  \cite{park2022majority}  &  52.08 & 55.41 & 63.65 & 84.71 & 86.80  & 91.18 \\
    ~ + ConCutMix (Ours)  &	\textbf{53.16} (\textbf{+1.15})&	\textbf{57.40} (\textbf{+1.08})	&\textbf{64.53 }(\textbf{+0.52}) & \textbf{86.07} (\textbf{+1.76})	&\textbf{88.00} (\textbf{+0.74})	&\textbf{91.42} (\textbf{+0.51})	
    
    \\
\hline
\end{tabular}
}
\end{center}
\end{table*}

\section{Experiment}
In this section, we evaluate the proposed ConCutMix in several benchmark long-tailed datasets including CIFAR-10-LT, CIFAR-100-LT, ImageNet-LT and iNaturalist 2018. We first describe the implementation details on each dataset in Section~\ref{sec:4.1}. Then, we compare ConCutMix with existing data augmentation methods for long-tailed recognition and other advanced methods in Section~\ref{sec:4.2}.

\subsection{Implementation Details}\label{sec:4.1}
\noindent\textbf{Long-Tailed CIFAR.}
CIFAR-10/100-LT follow an exponential decay across different
classes~\cite{cao2019learning}. They are created by reducing the number of training examples per class, but with the validation set unchanged. 
An imbalance ratio $r$ is used to denote the ratio between the number of samples of the most frequent and least frequent class, i.e., ${r = N_{max}/N_{min}}$. 
Following~\cite{zhu2022balanced}, we use the ResNet-32~\cite{he2016deep} with AutoAugment~\cite{cubuk2019autoaugment} as the backbone. 
The batch size is set to 256 and the weight decay is 5e-4 and the temperature $\tau$ for contrastive learning is set to 0.1.
{As ConCutMix is disabled by setting $\omega=0$ in the first 100/160 epochs for total 200/400 epochs.} 

\noindent\textbf{ImageNet-LT.}
{ImageNet-LT~\cite{liu2019large} is a long-tailed subset sampled from vanilla ImageNet~\cite{russakovsky2015imagenet} according to the Pareto distribution with a power value of 0.6.}
The batch size is set to 256 and we perform the same augmentation for branches as \cite{zhu2022balanced}.
We use ResNet-50~\cite{he2016deep} and ResNeXt-50-32x4d~\cite{xie2017aggregated} as the backbones and train them for 100 and 90 epochs.
These two backbones are trained with the same initial learning rate of 0.2 and weight decay of 1e-4, but different numbers of epochs, namely 100 and 90.
ConCutMix is disabled during the first 70 and 60 epochs for two backbones, respectively. 

\noindent\textbf{iNaturalist 2018.}
iNaturalist 2018  \cite{van2018inaturalist} is a large-scale long-tailed dataset containing 8,142 classes with 437,513 training and 24,424 validation images. 
We follow the official training and validation splits of iNaturalist 2018 in our experiments. 
We choose ResNet-50 as the backbone and follow almost the same training settings for 100 epochs as in ImageNet-LT.
{However, we adjust the batch size to 256 through gradient accumulation and enable ConCutMix during the final 20 epochs.}

\noindent\textbf{Places-LT.}
{
Places-LT is a long-tailed version of the large-scale scene classification dataset Places~\cite{zhou2017places}. It comprises 184.5K images spanning 365 categories, with the number of samples per category ranging from 5 to 4,980, exhibiting a significant long-tailed distribution. 
Following PaCo~\cite{cui2021parametric}, we employ ResNet-152 as the backbone network, pre-trained on the full ImageNet-2012 dataset (provided by torchvision), and fine-tune it for 30 epochs on Places-LT, which enable ConCutMix on last 20 epochs. The learning rate decays from 0.02 to 0 following a cosine scheduler with a batch size of 128. For balancing the weight of contrastive loss and CE loss , $\alpha$ is set to 0.7 and $\beta$ is set to 2.
}

\subsection{Comparisons with State-of-the-arts}\label{sec:4.2}

\noindent\textbf{Long-tailed CIFAR.}
In Table~\ref{cifar-table}, we report the comparison results between ConCutMix and existing advanced methods on CIFAR-10/100-LT.
ConCutMix consistently outperforms the other methods for considered imbalance factors including 100, 50, and 10.
Recent data augmentation methods for long-tailed recognition \cite{chou2020remix,xu2021towards,jeong2022supervised} 
show limited performance because they ignore the semantic information of synthetic samples.
For a fair comparison, we apply some pluggable augmentation methods to baseline BCL~\cite{zhu2022balanced} including Mixup~\cite{zhang2017mixup}, CutMix~\cite{yun2019CutMix} and CMO~\cite{park2022majority}, whose training curves of accuracy are plotted in Fig.~\ref{fig:acc}.
We observed that the considered pluggable augmentation methods do not always improve the baseline, while ConCutMix brings significant accuracy gap to the baseline.
{Besides, we compare the accuracy of BCL~\cite{zhu2022balanced} and ConCutMix at different imbalance factors in Fig.~\ref{fig:gap}. We highlight that ConCutMix can achieve limited improvement on balanced tasks but becomes more effective as the degree of imbalance rises (i.e., with a larger imbalance factor).
{
More critically, in addition to the very recent and strong baseline BCL~\cite{zhu2022balanced}, our ConCutMix can easily generalize to diverse baseline methods and yield significant improvements as shown in Table~\ref{tab:baselines}.
This table comprehensively compares the performance gains achieved by applying ConCutMix to various state-of-the-art long-tailed classification baselines (e.g., LDAM-DRW, BBN, ResLT, SBCL, SHIKE, PaCo, GPaCo).
For example, our ConCutMix contributes to a large improvement of approximately 5\% in performance for PaCo at 100 imbalance factor. Similar notable gains are also observed for the generalized GPaCo.}
We argue that the results show a conflict between the boost from expanding tail classes and the hindrance from noisy training signals, which is effectively resolved by ConCutMix.

\begin{figure}[t]
    \centering
    \centerline{\includegraphics[width=0.81\columnwidth]{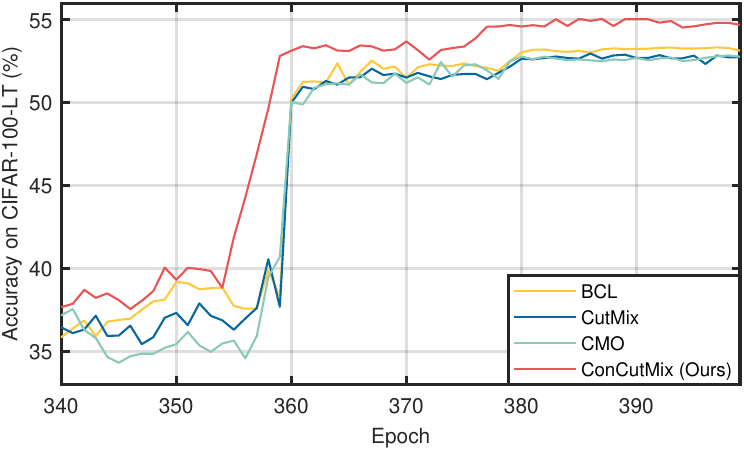}}
    \caption{Comparisons of the training curves of accuracy among different methods implemented with ResNet-32 on CIFAR-100-LT at 100 imbalance factor for 400 epochs. ConCutMix consistently outperforms other methods during last 60 epochs.
    }
    \label{fig:acc}
\end{figure}

\begin{figure}[t]
    \centering
    \centerline{\includegraphics[width=0.9\columnwidth]{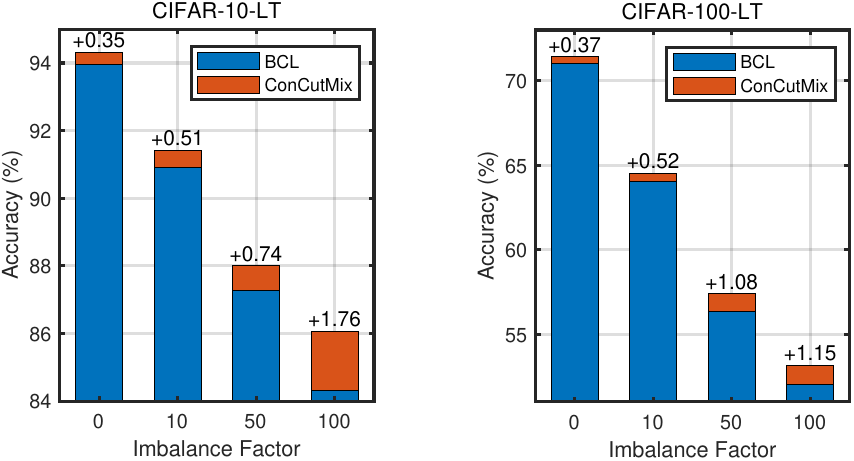}}
    \caption{
        {Comparisons of Top-1 Accuracy between ConCutMix and baseline BCL as different imbalance factors. A value of 0 for the imbalance factor indicates the models trained on vanilla CIFAR. ConCutMix brings larger improvements as the degree of imbalance rises.}
    }
    \label{fig:gap}
\end{figure}

\begin{table*}[ht]
\begin{center}
\caption{{Comparisons of Top-1 accuracy of applying ConCutMix to various baseline on CIFAR-100-LT and CIFAR-10-LT for 200 epochs. ConCutMix can easily generalize to different long-tailed baseline methods and bring significant improvements in performance.}}\label{tab:baselines}
\resizebox{\linewidth}{!}{
\begin{tabular}{l|ccc|ccc}
\hline
  Method &  \multicolumn{3}{c|}{CIFAR-10-LT} & \multicolumn{3}{c}{CIFAR-100-LT}\\
  \hline
  Imbalance Factor & 100 & 50 & 10 & 100 & 50 & 10\\ \hline
  \multicolumn{7}{c}{ ResNet-32} \\ \hline
  LDAM-DRW~\cite{cao2019learning}   & 77.03 & 81.03 & 88.16 & 42.04 & 46.62 & 58.71\\
  ~+ ConCutMix  & \textbf{79.28} (\textbf{+2.25})& \textbf{82.24} (\textbf{+1.21})&\textbf{88.44} (\textbf{+0.28})& \textbf{45.66} (\textbf{+3.62})& \textbf{49.06} (\textbf{+2.44})&\textbf{59.17} (\textbf{+0.46})\\
  \hline
  
  BBN~\cite{zhou2020bbn}  & 79.82 & 81.18 & 88.32 & 42.56 & 47.02 & 59.12 \\
  ~+ ConCutMix  & \textbf{80.69} (\textbf{+0.87})& \textbf{84.09} (\textbf{+2.91})& \textbf{89.26} (\textbf{+0.94})&  \textbf{44.45} (\textbf{+1.89})& \textbf{47.97} (\textbf{+0.95})& \textbf{60.16} (\textbf{+1.04})\\
  \hline
  ResLT~\cite{cui2022reslt}  & 79.54	&83.18	&88.77 &45.37&	49.99	&59.56 \\ 
  ~+ ConCutMix &\textbf{81.92} (\textbf{+2.38})& \textbf{84.19} (\textbf{+1.01})& \textbf{89.03} (\textbf{+0.26})& \textbf{49.74} (\textbf{+4.37})& \textbf{53.34} (\textbf{+3.35})&\textbf{62.25} (\textbf{+2.69})\\  
  \hline
    SBCL~\cite{hou2023subclass} &	74.92	&80.01	&84.51  & 44.90	&48.70	&57.90
 \\  
~+ ConCutMix &\textbf{76.41 }(\textbf{+1.49})	&\textbf{81.21} (\textbf{+1.20})	&\textbf{84.92 }(\textbf{+0.41})& \textbf{46.81 }(\textbf{+1.91})	&\textbf{49.10 }(\textbf{+0.40})	&\textbf{58.41 }(\textbf{+0.51})
\\\hline
SHIKE~\cite{jin2023long} &84.40	&87.57&	91.43& 55.12&	61.13&	67.04	

 \\ 
  ~+ ConCutMix &	\textbf{85.21}	 (\textbf{+1.12})&\textbf{87.82}	 (\textbf{+0.25})&\textbf{91.62 }(\textbf{+0.19})&\textbf{ 55.64} (\textbf{+0.52})&	\textbf{61.51} (\textbf{+0.38})	&\textbf{67.41} (\textbf{+0.37})	\\
  \hline
  \multicolumn{7}{c}{ ResNet-50} \\ \hline
PaCo~\cite{cui2021parametric}   & 76.10 & 85.77 & 91.27 & 52.00& 56.00 & 64.20 \\ 
  ~+ ConCutMix & \textbf{81.03} (\textbf{+4.93})& \textbf{87.05} (\textbf{+1.28})& \textbf{91.43} (\textbf{+0.16})&\textbf{53.41} (\textbf{+1.41})& \textbf{56.89} (\textbf{+0.89})& \textbf{64.68} (\textbf{+0.48})\\
  \hline
GPaco~\cite{cui2023generalized} &   	76.11&	85.77&	91.27&52.30&	56.40&	65.40
 \\ 
  ~+ ConCutMix & \textbf{81.07}	(\textbf{+3.84})&\textbf{87.34 }(\textbf{+1.57})	&\textbf{91.51} (\textbf{+0.24})&\textbf{53.61 }(\textbf{+1.31})& 	\textbf{57.02} (\textbf{+0.62)}&	\textbf{65.92 }(\textbf{+0.52})
\\
  \hline
\end{tabular}}
\end{center}
\end{table*}

\noindent\textbf{ImageNet-LT and iNaturalist 2018.}
Table \ref{ima_ina-table}  lists the results of different methods implemented with ResNet-50 on ImageNet-LT and iNaturalist 2018, both of which are large scale long-tailed datasets.
LWS  \cite{kang2019decoupling}, $\tau$-norm  \cite{kang2019decoupling}, and DisAlign  \cite{zhang2021distribution} adopt the two-stage learning strategy and focus on fine-tuning the classifier in the second stage, but neglecting the data scarcity the tail class.
Hybrid-SC/-PSC \cite{wu2021contrastive}, TSC \cite{li2022targeted} and BCL~\cite{zhu2022balanced} using a contrastive learning branch and a classifier branch for eliminating the bias of the classifier. But the samples used for the classifier in these methods are severely unbalanced.
Augmentation methods such as Mixup~\cite{zhang2017mixup}, CutMix~\cite{yun2019CutMix} and CMO~\cite{park2022majority} show limited improvements over baseline, because they produce much noisy training signals on these large scale datasets.
By contrast, ConCutMix exceeds BCL~\cite{zhu2022balanced} 2.6\% on accuracy on ImageNet-LT, and achieves the overall accuracy at 72.1\% on iNaturalist 2018. {Besides, the results in Table~\ref{table: ina-400-rebuttal} and Table~\ref{90-table-rebuttal} shows the proposed ConCutMix still works well on various backbones, including ResNet-50, ResNet152, ResNeXt-50, ResNeXt-101.} 

\begin{table}
\caption{
Comparisons of Top-1 accuracy of different methods implemented with ResNet-50 on ImageNet-LT and iNaturalist18 for 100 epochs.
ConCutMix shows better performances on these large-scale long-tailed datasets.
The values in parentheses show the accuracy improvements
of ConCutMix over BCL.
The best results are marked in bold.}
\begin{center}
\resizebox{0.9\linewidth}{!}{
\begin{tabular}{l|cccc}
\hline
Method & ImageNet-LT& iNaturalist 2018\\
\hline
  $\tau$-norm  \cite{kang2019decoupling} & 46.7  & 65.6 \\
    cRT  \cite{kang2019decoupling}  & 49.6  & 65.2 \\
    LWS  \cite{kang2019decoupling}  & 49.9  & 65.9 \\
    BBN  \cite{zhou2020bbn}  & -     & 66.3 \\

    SSP  \cite{yang2020rethinking}  & 51.3  & 68.1 \\

    RIDE (2 experts) \cite{wang2020long} & 54.4  & 71.4 \\
    KCL \cite{kang2020exploring} &   52.4  & 68.6\\
    Hybrid-SC  \cite{wang2021contrastive} & -     & 66.7 \\
    Hybrid-PSC  \cite{wang2021contrastive} & -     & 68.1 \\
    SMC \cite{jeong2022supervised} &56.6 &70.6\\   

    UniMix \cite{xu2021towards} & 48.4 &  - \\   
    Disalign  \cite{zhang2021distribution} & 52.9  & 69.5 \\
    TSC \cite{li2022targeted}& 52.4 & 69.7\\
    ResLT \cite{cui2022reslt} & - & 70.2 \\

    \hline
    
    BCL  \cite{zhu2022balanced}  & 55.9    & 71.1 \\

    ~ + Mixup \cite{zhang2017mixup} & 56.3& 67.9\\
    ~ + CutMix \cite{yun2019CutMix} &57.1 &67.8\\
    ~ + CMO  \cite{park2022majority}  & 57.3 &69.3 \\

    ~ + ConCutMix (Ours)  &  \textbf{58.5 (+2.6)}  & \textbf{72.1 (+1.0)}   \\
\hline
\end{tabular}}
\end{center}
\label{ima_ina-table}
\end{table}

\begin{table}[t]
\centering

    \caption{{Comparisons of Top-1 accuracy of different methods implemented with various backbone on iNaturalist18 for 400 epochs.
ConCutMix shows better performances with various backbones for longer training epochs. The best results are marked in bold.}}
    \resizebox{1\columnwidth}{!}{
    \begin{tabular}{l|cccc}
    \hline
     Method  & Many & Medium & Few &All \\
    \hline
    ResNet-50
    \\
    \hline
    PaCo~\cite{cui2021parametric}	&{70.3}	&{73.2}	&{73.6}&	{73.2}\\
    GPaCo~\cite{cui2023generalized}	 & - & 	- & 	-	 & {75.4}
    \\
        
        BCL~\cite{zhu2022balanced} & 71.4	&{73.6}	&{74.9}	&{74.0}
    \\
    
        ~+ ConCutMix& \textbf{76.1 (+4.7)}	&\textbf{75.9 (+2.3)}&	\textbf{76.2 (+1.3)}&	\textbf{76.0 (+2.0)}
      \\
        \hline
    ResNet-152\\
    
       \hline
    PaCo~\cite{cui2021parametric}	&{75.0}&	{75.5}&	{74.7}	&{75.2}\\
    GPaCo~\cite{cui2023generalized}	 & - & 	- & 	-	 & {78.1}\\
    
        BCL~\cite{zhu2022balanced}   & {78.4}	&{77.2}	&{76.3}	&{76.8}
     \\
        ~+ ConCutMix&  \textbf{79.4 (+1.0)} &\textbf{78.5 (+1.3)}	& \textbf{78.3 (+2.0)}	& \textbf{78.5 (+1.7)}
      \\
        \hline 
    \end{tabular}}
    \label{table: ina-400-rebuttal}
\end{table}

\begin{table}[t]
\caption{{Comparisons of Top-1 accuracy of different methods implemented with various backbone on ImageNet-LT for 400 epochs.
ConCutMix shows better performances with various backbones for longer training epochs. The best results are marked in bold.}}\label{90-table-rebuttal}
\begin{center}

\resizebox{1\columnwidth}{!}{
\begin{tabular}{l|cccc}
\hline
 Method  & Many & Medium & Few &All \\

\hline

ResNet-50  \\
\hline

PaCo~\cite{cui2021parametric} &65.0	&55.7	&38.2&	57.0\\
GPaCo~\cite{cui2023generalized}&	-&-&-&		58.5\\
BCL~\cite{zhu2022balanced} &66.1	&55.9&	37.1&57.2\\
~+ Mixup~\cite{zhang2017mixup}  &66.1	&56.3	&37.4	&57.5
\\
~+ CutMix~\cite{yun2019CutMix}  &66.4&	56.5	&38.1	&57.8
\\
~+ CMO~\cite{park2022majority}  &63.4	&55.6	&\textbf{41.1}&	56.6
\\
~+ ConCutMix (Ours)&\textbf{67.5} \textbf{(+1.4)}&	\textbf{57.9} \textbf{(+2.0)}	&37.9	 (+0.8)&\textbf{58.9} \textbf{(+1.7)}\\

\hline
ResNeXt-50  \\
\hline
PaCo~\cite{cui2021parametric} & 67.5	&56.9	&36.7&	58.2\\
GPaCo~\cite{cui2023generalized}&	-& -& -&		58.9\\
BCL~\cite{zhu2022balanced} & 67.8	&56.8	&36.9	&58.3\\

~+ Mixup~\cite{zhang2017mixup}  &67.8	&57.2&	37.0	&58.5

\\
~+ CutMix~\cite{yun2019CutMix}  &68.4	&57.5	&37.1&	58.9

\\
~+ CMO~\cite{park2022majority}  &65.2&	56.4	&\textbf{42.9}	&57.9

\\
 ~+ ConCutMix (Ours)&  \textbf{72.1} \textbf{(+4.3)}	&\textbf{58.4} \textbf{(+1.6)}&	40.8 (+3.9)	&\textbf{61.3} \textbf{(+3.0)}\\

\hline
ResNeXt-101  \\
\hline
PaCo~\cite{cui2021parametric}&	 68.2&	58.7	&41.0	&60.0\\
GPaCo~\cite{cui2023generalized}	 & -& 	-& -	 & 60.8\\

BCL~\cite{zhu2022balanced}    & 67.4& 	56.9	& 40.2	& 58.6
\\
~+ Mixup~\cite{zhang2017mixup}  &67.4& 	56.9& 	40.8	& 58.7
\\
~+ CutMix~\cite{yun2019CutMix}  &67.7& 	57.3	& 40.9 & 	59.0
\\
~+ CMO~\cite{park2022majority}  & 64.4	& 56.1& 	\textbf {45.8}	& 57.8
\\
~+ ConCutMix (Ours)& \textbf{72.1}  \textbf{(+3.7)}&	\textbf{59.4}  \textbf{(+2.5)}&	41.1 (+0.9)	&\textbf {61.8}  \textbf{(+3.2)}

\\
    \hline
\end{tabular}}
\end{center}
\end{table}

{
\noindent\textbf{Places-LT.}
Table~\ref{plt-30} shows the results on the Places-LT dataset after fine-tuning for 30 epochs. Compared to baseline methods like PaCo~\cite{cui2021parametric}, GPaCo~\cite{cui2023generalized}, and BCL~\cite{zhu2022balanced}, ConCutMix achieves consistent performance gains across different groups of classes, especially for the Few-shot classes. Specifically, when combined with the BCL baseline, ConCutMix brings an accuracy gain of 1.7\% for Many-shot, 0.7\% for Medium-shot, and a notable 7.0\% for Few-shot classes, resulting in an overall 2.3\% improvement.
}

\begin{table}[ht]
\caption{
{Comparisons of Top-1 accuracy of different methods implemented with ResNet-152 on Places-LT for 30 epochs.
ConCutMix surpasses other data augmentation methods and baseline methods, achieving consistent and substantial gains across different groups of classes, with remarkable improvements for the Few-shot classes.
}}\label{plt-30}
\begin{center}

\resizebox{1\linewidth}{!}{
\begin{tabular}{l|cccc}
    \hline
     Method  & Many & Medium & Few &All \\
    \hline
    PaCo~\cite{cui2021parametric}	 & 37.5 & 	47.2 &   33.9	  & 41.2\\
    ~+ ConCutMix (Ours)&  \textbf{38.4} (\textbf{+0.9})	&\textbf{48.2} (\textbf{+1.0})&	\textbf{35.1} (\textbf{+1.2})	& \textbf{42.1} (\textbf{+0.9})
       \\
    \hline
    
        GPaCo~\cite{cui2023generalized} & 39.5  &47.2  &33.0  & 41.7
     \\
        
        ~+ ConCutMix (Ours)  & \textbf{39.9} (\textbf{+0.4}) &\textbf{47.8} (\textbf{+0.6})  &\textbf{34.9} (\textbf{+1.9})  & \textbf{42.2} (\textbf{+0.5}) \\
        \hline
        BCL~\cite{zhu2022balanced}   & 43.8& 39.2& 22.9 & 37.7
     \\
        ~+ Mixup \cite{zhang2017mixup} &45.0&39.1& 22.5 &38.1\\
        ~+ CutMix \cite{yun2019CutMix} &45.9
     &39.6
    &21.9
     &38.5\\
        ~+ CMO \cite{park2022majority} & 35.9& 39.6&41.9 &38.8\\
        ~+ ConCutMix (Ours)&  \textbf{45.5} (\textbf{+1.7})  &  \textbf{39.9} (\textbf{+0.7})  &  \textbf{29.9} (\textbf{+7.0})   &  \textbf{40.0} (\textbf{+2.3})   \\
        \hline
\end{tabular}}
\end{center}
\end{table}

{
\noindent\textbf{Comparisions with MoE.} MoE (Mixture of Experts)
            methods are renowned in long-tailed recognition for their superior performance. In Table~\ref{cifar-table} to Table~\ref{ima_ina-table}, we have compared representative methods from recent years, including BBN~\cite{zhou2020bbn}, RIDE~\cite{wang2020long}, ResLT~\cite{cui2022reslt}, and SHIKE~\cite{jin2023long} (CVPR 2023). Experimental results show that advanced single-model methods can outperform MoE methods with fewer experts in terms of performance. However, the recent SHIKE significantly outperforms single-model methods like BCL on complex datasets such as ImageNet and iNaturalist2018.
            Notably, MoEs can achieve larger performance gains by increasing the number of experts and are frequently used in practice. However, MoEs typically require greater computational resources. For instance, training SHIKE + ConCutMix on CIFAR-LT is about 20\% slower than training BCL + ConCutMix. The effectiveness of MoEs also hinges on carefully designed architectures and training workflows to cultivate a diversity of experts, which often makes integration with other methods challenging. In contrast, ConCutMix, a data augmentation method, focuses on leveraging existing data to generate additional information and is mostly independent of specific model structures and training processes. Therefore, it can be easily integrated into other methods, including certain MoEs. MoEs explicitly enables different experts to learn from classes of varying frequencies without sacrificing the accuracy of head classes, while ConCutMix meticulously adjusts the confidence in semantically consistent labels to effectively bolster head classes and facilitate knowledge transfer between head and tail classes.
}

\subsection{Ablation Study}
\label{discussion}
In this section, we conduct extensive experiments and further discussions about the components and hyperparameters of ConCutMix to illustrate its effectiveness.

\noindent\textbf{Similarity Measurement.}
In a discriminative feature space constructed by contrastive learning, two popular metrics for measuring the similarity between two features are Euclidean distance~\cite{yang2011spectral} and cosine similarity  \cite{cevikalp2022deep}. 
We report the influence of these two similarity measurements for ConCutMix in Table \ref{tab: weighting method}. 
The experimental results show ConCutMix with different similarity measurements are always better than the baseline.
Therefore, we empirically use Euclidean distance to compute similarities in other experiments.

\begin{table}[t]
\caption{Comparisons of accuracy of different similarity measurements on CIFAR-100-LT at 100 imbalanced factor for 400 epochs and iNaturalist18 for 50 epochs.
ConCutMix with different similarity measurements are better than BCL.}\label{tab: weighting method}
\begin{center}
\resizebox{\linewidth}{!}{

\begin{tabular}{l|cc}
\hline
       Method  & CIFAR-100-LT  & iNaturalist 2018 \\
\hline

    BCL \cite{zhu2022balanced} (baseline) & 53.4 & 67.48 \\
    ConCutMix (Cosine Similarity) & 54.5 (+1.1) & 67.72 (+0.24) \\
    ConCutMix (Euclidean Distance) & \textbf{55.1} (\textbf{+1.7}) & \textbf{68.30} (\textbf{+0.82}) \\
\hline
\end{tabular}
}
\end{center}
\end{table}

\noindent\textbf{Number of Associated Classes $\mathrm{K}$.}
A sample synthesized by CutMix could possibly belong to a novel class that is not included in the corresponding area-based label.
To address this issue, ConCutMix considers TopK-similar classes for each synthetic sample, where the two classes used for CutMix are always included.
The results of ConCutMix with different $\mathrm{K}$ are shown in Fig.~\ref{ablation} (left), where ConCutMix is always better than BCL by a notable margin.
We noticed that it shows limited improvement if extra semantically similar classes are not considered, i.e., $\mathrm{K}=2$. 
Besides, as $\mathrm{K}$ increases, the performance of ConCutMix may degrade, because introducing some classes to which synthetic samples are almost impossible to belong to may hinder the training.
We empirically set $\mathrm{K}=0.3\mathcal{|Y|}$, but the maximum does not exceed 30.

\noindent\textbf{Confidence Function $\gamma(\cdot)$.}
We focus on scaling factors $\omega$ and activation function~$\varphi$ to generate different confidence functions $\gamma(\cdot)$. 
As is shown in Figure~\ref{ablation} (right), ConCutMix with different $\omega$ has certain improvements compared to baseline.
{The performance does not change with apparent regularity when the $\omega$ is changing, so we empirically set the $\omega$ for different datasets. We hypothesize a linear relationship, namely $w=k|Y|$, where $k$ is a hyperparameter, $|Y|$ represents the number of classes in the dataset. In our experiments, the value of $w$ for different datasets is obtained by setting $k$ to the same value of 8e-4.
}
As for the activation function, we use the linear function and logarithmic function to generate two confidence functions denoted as $\gamma$-${linear}$ and $\gamma$-${log}$, respectively. 
No matter what activation function is used in the confidence function, ConCutMix always significantly boosts the baseline as shown in Table~\ref{tab: function}.
In particular, when $\gamma=0.5$, i.e., semantically consistent labels and area-based labels are treated equally, ConCutMix can still rectify the noisy training signals to achieve better performance.

\begin{figure}[t]
    \centering
    \vspace{1 pt}
    \centerline{\includegraphics[width=1\columnwidth]{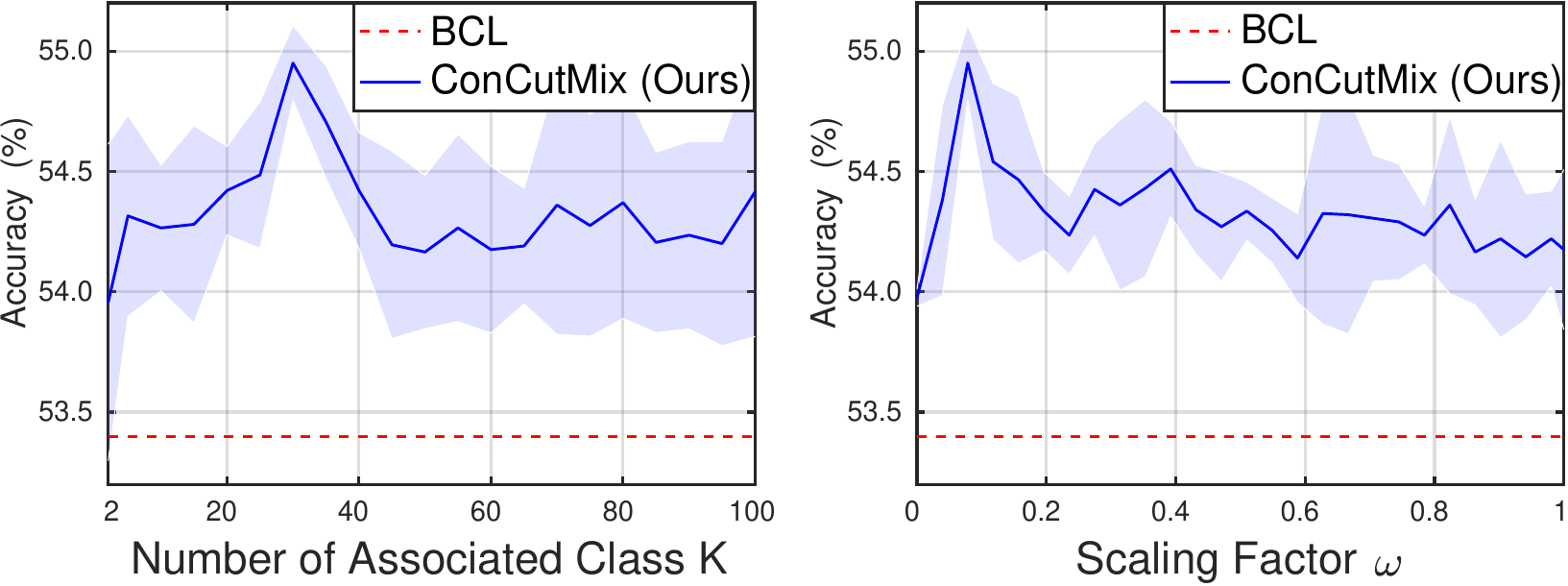}}
    \vspace{7 pt}
    \caption{
Comparison of accuracy of different numbers of associated classes K~(left) and scaling factors~(right)  with ResNet-32 on CIFAR-100-LT at 100 imbalance factor for 400 epochs.
$\mathrm{K}=2$ means the classes not used for CutMix would not be considered. $\omega=0$ means to use CutMix but not ConCutMix.
Only using CutMix or using ConCutMix with two classes shows limited improvements over BCL.
}
\label{ablation}
\end{figure}

\begin{table}[t]
\caption{Comparison of accuracy of different designs of confidence function on CIFAR-100-LT at 100 imbalance factor for 400 epochs and iNaturalist18 for 50 epochs.
A constant value of $\gamma$ means to disable dynamic confidence for different synthetic samples.
$\gamma$-${linear}$ and $\gamma$-${log}$ denote confidence functions using linear and logarithmic activation functions, respectively.
ConCutMix always boosts BCL \cite{zhu2022balanced} and achieves higher performances with well-designed confidence functions.
}\label{tab: function}
\begin{center}
\resizebox{\linewidth}{!}{
    \begin{tabular}{l|cccc}
\hline
      Method & BCL \cite{zhu2022balanced} &   ConCutMix $(\gamma$=0.5) & ConCutMix ($\gamma$-${linear}$) & ConCutMix ($\gamma$-${log}$)\\
\hline
     CIFAR-100-LT & 53.4 & 54.1 (+0.7)   & 54.5 (+1.1) & \textbf{55.1} (\textbf{+1.7})  \\
     iNaturalist18 & 67.48 & 67.53 (+0.05) & 67.74 (+0.26) & \textbf{68.30} (\textbf{+0.82}) \\
\hline
\end{tabular}
}
\end{center}

\end{table}

\begin{table}[t]
\caption{Comparisons of the different combinations of samplers with ResNet-32 on CIFAR-100-LT at 100 imbalance factor for 400 epochs. BCL simply uses a random sampler to sample data. ConCutMix benefits from the random background sampler and the foreground sampler, yielding the better performance.}\label{tab: sampler method}
\begin{center}
\resizebox{0.9\linewidth}{!}{
\begin{tabular}{l|ccc}
\hline
 Method & \makecell[c]{Background \\ Sampler} & \makecell[c]{Foreground \\ Sampler} 
    & Acc. (\%) \\
\hline
  {BCL \cite{zhu2022balanced} (baseline)} &-&-  & 53.4 \\ \hline
    \multirow{5}{*}{ConCutMix} & Balanced & Random & 53.2\\
     &Reversed & Random  &48.6\\
    & Random &  Random  & 53.4 \\
    & Random & Reversed& 53.5  \\
     &Random & Balanced  &\textbf{55.1} (\textbf{+1.7}) \\
\hline
\end{tabular}
}
\end{center}
\end{table}

\noindent\textbf{Combination of Samplers.}
ConCutMix uses two samplers to get images for synthesizing samples.
We consider the combinations of balanced \cite{shen2016relay}, random, and reversed \cite{zhou2020bbn} samplers to explore the effect of samplers on ConCutMix.
As shown in Table~\ref{tab: sampler method}, we found that sampling the background and foreground using a random sampler and a balanced sampler, respectively, can significantly improve the model performance.
We believe that sampling the foreground with a balanced sampler allows for more synthetic samples dominated by tail classes without underfitting on head classes.
Besides, sampling the background more from the head classes with a random sampler can benefit the model, which is similar to the conclusion of CMO \cite{park2022majority}. 
However, we simply use these samplers in ConCutMix, without carefully adjusting the data distribution like CMO \cite{park2022majority}.

\noindent \textbf{Combining Area-based and Semantically Consistent Label.}
{
It is worth emphasizing that an inaccurate model in the early stage could affect the performance, i.e., hampering the quality of semantically consistent labels.
Specifically, non-representative class centers are constructed from a noisy semantic space, leading to low-quality semantically consistent labels.
To mitigate this issue, as mentioned in the implementation details, we disable ConCutMix ($\omega=0$) in the early stage and instead train the model solely using area-based labels. In this way, we can construct better semantically consistent labels from representative class centers. 
Subsequently, we combine  the constructed semantically consistent labels and area-based labels to rectify noisy training signals.
As shown in Table~\ref{tab: semantic label alone}, combining both labels (Ours) significantly outperforms area-based labels, which verifies our effectiveness.
In the later stages, combining both labels significantly outperforms replacing area-based labels with semantically consistent labels, due to the quality of semantically consistent labels can still benefit from the ground truth information of area-based labels.
Besides, we observe a severe accuracy drop when training with semantically consistent labels alone. The main reason is that, without area-based labels, we cannot receive any supervision signal (i.e., fully unsupervised) and semantically consistent labels become very noisy.}

\setlength{\tabcolsep}{16pt} 
\begin{table}[t]
\caption{\centering {Comparisons of accuracy of different combinations of area-based label and semantically consistent label for training on CIFAR-LT at 100 imbalance factor for 200 epochs. ConCutMix rectifies area-based labels with semantically consistent labels, resulting in significantly improved performance compared to other label combinations.}}\label{tab: semantic label alone}
\begin{center}
\resizebox{1\linewidth}{!}{

\begin{tabular}{l|cccc}
\hline
  Training Labels & CIFAR-10-LT & CIFAR-100-LT\\
  \hline
Area-based  & 84.31 & 52.01  \\
Semantic & 18.01 & 15.60    \\
Area then Semantic & 84.11 & 51.90 \\
Area + Semantic (Ours) & \textbf{86.07} (\textbf{+1.76}) & \textbf{53.16} (\textbf{+1.15}) \\
\hline
\end{tabular}
}
\end{center}
\end{table}

\begin{table}[ht]
\caption{Comparisons of accuracy on CIFAR-100-LT with different combinations of data augmentation types and number of training batches for 400 epochs. Increasing the number of batches alone yields marginal improvements, while combining the increased number of batches with the proposed ConCutMix method achieves significant performance boosts over the baseline.}\label{table:batches}
\begin{center}

\resizebox{1\columnwidth}{!}{
\begin{tabular}{l|c|cc}
\hline
 Method &  Batch Num  & Acc. (\%) &Time (Mins) \\
\hline
\multirow{3}{*}{CE}  & 1  &	51.2  &51\\ 

  & 2     &51.9  &57\\ 

  & 3     & 52.7 &63  \\
  \hline

BCL~\cite{zhu2022balanced} & \multirow{2}{*}{3} &53.4   &58\\ 

 + ConCutMix &  & \textbf{55.1} (\textbf{+1.7})&60
\\

    \hline 
\end{tabular}}
\end{center}
\end{table}

\noindent{\textbf{Training data batches. }
As demonstrated in Table~\ref{table:batches}, we explored the impact of varying the number of training batches on CIFAR-100-LT for 400 epochs. While simply increasing the number of cross-entropy (CE) batches from 1 to 3 yields marginal improvements in accuracy from 51.2\% to 52.7\%, combining the same three batches with our proposed ConCutMix method achieves a significant boost to 55.1\%, outperforming the BCL by 1.7\%. These results validate that the primary performance gain of ConCutMix does not stem from utilizing more training data, but rather from the effective data augmentation and the construction of semantically consistent labels. Furthermore, Table~\ref{table:batches} shows that increasing the number of batches does not significantly increase the training time overhead due to our efficient parallelization of multiple batch processing.
    }
}

\begin{figure*}[h]
    \centering
    \centerline{\includegraphics[width=\linewidth]{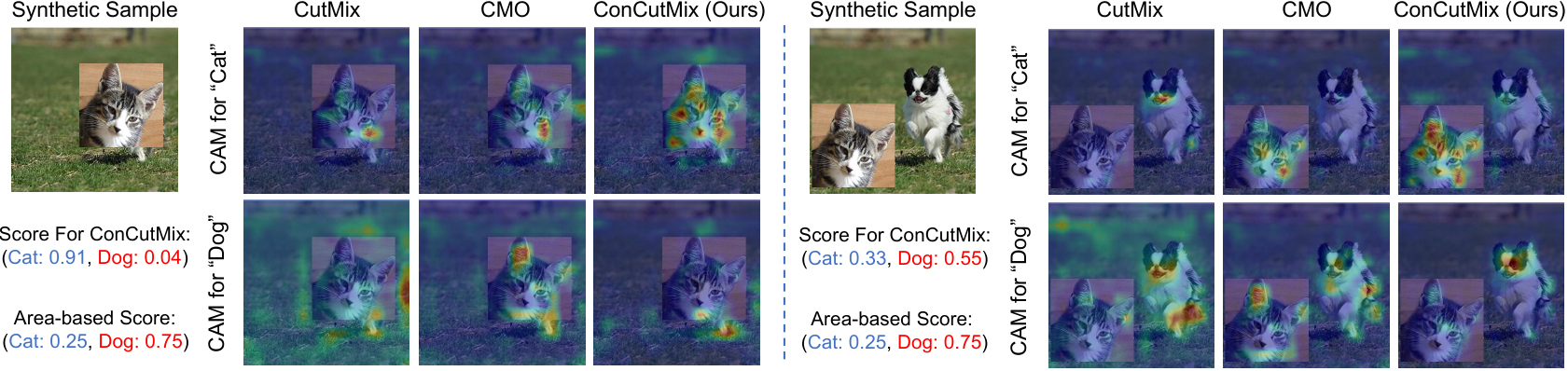}}
    \caption{ 
    Visualization of class activation mapping (CAM)~\cite{zhou2016learning} for synthetic samples on the class ``Cat'' and ``Dog''. The considered synthetic samples are generated by CutMix~\cite{yun2019CutMix} from two images with the same area-based labels.
    For the class ``Cat'', ConCutMix pays more attention to the foreground which belongs to ``Cat'', which illustrates ConCutMix effectively captures the information of the synthetic sample, without being misled by noisy training signals from the area-based label. For the class ``Dog'', ConCutMix focus on the important feature of the dog, e.g., face and feet, instead of grass in the background or the cat in the foreground.}
    \label{fig:cam}
\end{figure*}

\begin{figure*}[ht]
    \centering
\resizebox{\linewidth}{!}{
    \begin{minipage}[t]{0.25\linewidth}
        \centering
        \includegraphics[width=\textwidth]{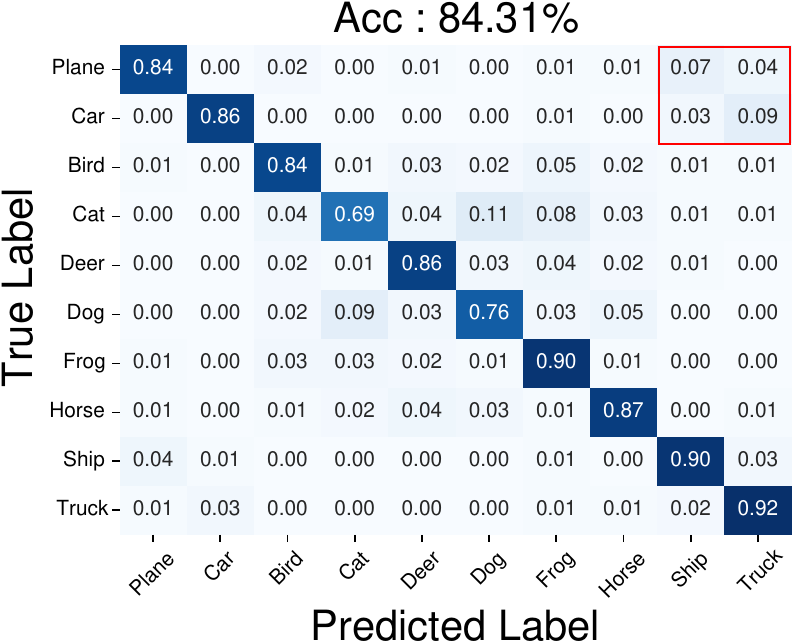}
        \centerline{ \quad BCL}
    \end{minipage}
    \begin{minipage}[t]{0.25\linewidth}
        \centering
        \includegraphics[width=\textwidth]{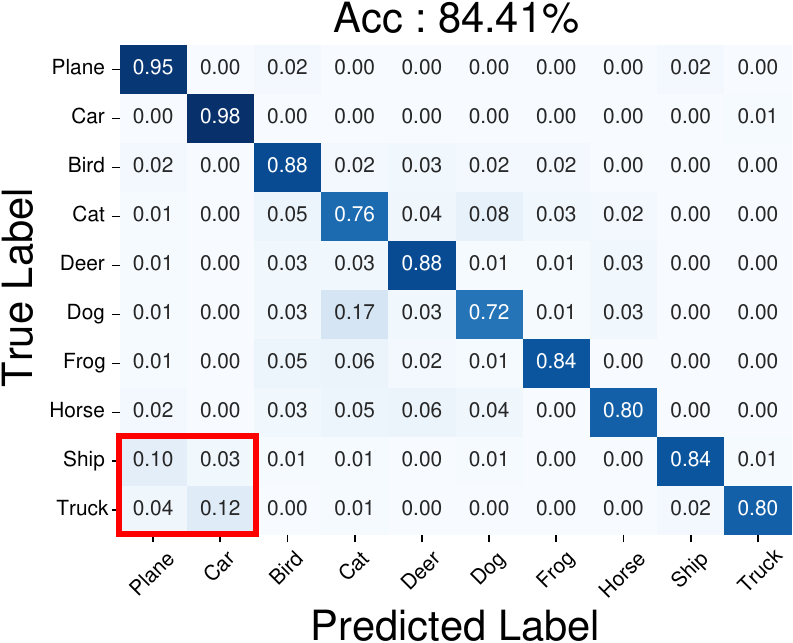}
        \centerline{ \quad CutMix}
    \end{minipage}
    \begin{minipage}[t]{0.25\linewidth}
        \centering
        \includegraphics[width=\textwidth]{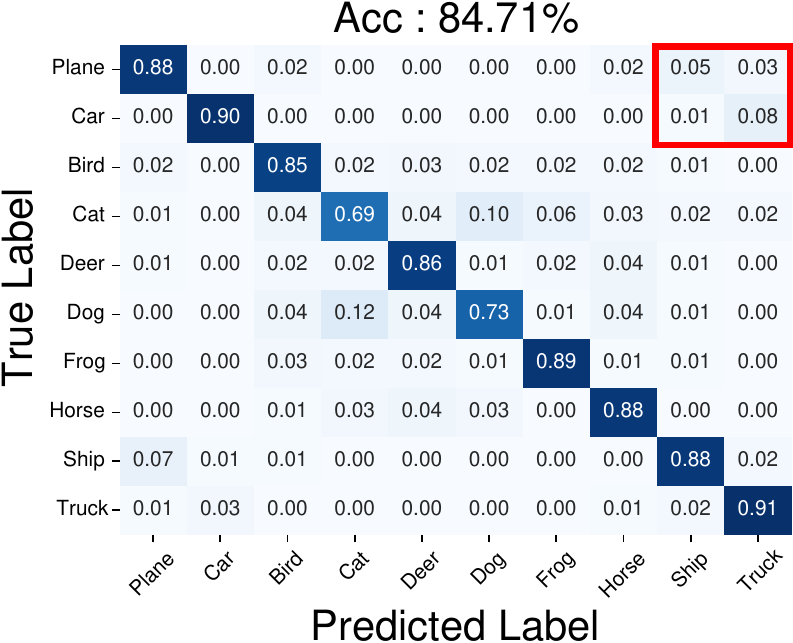}
        \centerline{ \quad CMO}
    \end{minipage}
        \begin{minipage}[t]{0.25\linewidth}
        \centering
        \includegraphics[width=\textwidth]{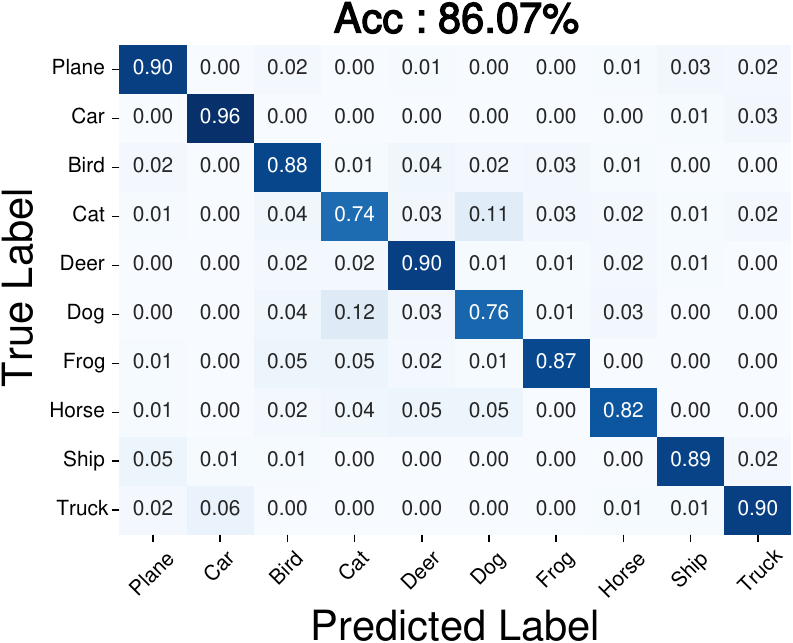}
        \centerline{ \quad ConCutMix (Ours)}
    \end{minipage}
    }

    \caption{Comparisons of the confusion matrices of prediction results among different methods implemented with ResNet-32 on CIFAR-10-LT at 100 imbalance factor for 200 epochs. The number of samples of classes decreases from the Plane to the Truck. Our ConCutMix significantly improves accuracy on the tail classes without sacrificing the accuracy on the head classes.
    We use visual annotations in the form of \red{red} boxes to emphasize the significant regions of confusion observed in each methods. Notably, ConCutMix reduces the confusion between head and tail classes.
    }
    \label{fig:confusion matrix}
\end{figure*}

\section{Further Discussions}
In this section, we explore the superiority of ConCutMix in long-tail recognition tasks from additional perspectives.

\noindent\textbf{Better Semantic Information from Synthetic Samples.}
{
To better understand the effectiveness of the ConCutMix in capturing semantic information, we employed the class activation mapping (CAM)~\cite{zhou2016learning} to identify the critical image regions for discriminating specific classes.
As depicted in Fig.~\ref{fig:cam}, we visualize the CAM of two synthetic samples with the same area-based label but different semantics.
For the left synthetic sample, its area-based label is entirely wrong intuitively due to the cat patch covering the dog body. In contrast, we find that ConCutMix, benefiting from semantically consistent labels, consistently pays more attention to features of cats in the foreground to identify the class ``Cat'' than other methods. 
As for the class ``Dog'', when most of the semantic information of dog is lost, ConCutMix can still capture the remaining semantic information in the synthetic sample, such as dog feet.
However, the noisy training signals make CutMix~\cite{yun2019CutMix} focus on the grass in the background and CMO~\cite{park2022majority} focus on cat features, both of which are unreasonable to be associated with the class ``Dog''.
These observations can also be drawn from the right synthetic sample, whose area-based label produces less noisy training signals.
In conclusion, no matter how we combine the images, ConCutMix can effectively capture semantic information of synthetic samples to rectify the noisy training signals produced by area-based labels.}

\noindent\textbf{Mitigating Spurious Correlation.}
{
Data augmentation methods often raise concerns about ``spurious correlations", where models erroneously rely on irrelevant patterns, such as background or texture, for the purpose of classification \cite{nagarajan2020understanding,sagawa2020investigation}.
This phenomenon arises from the potential absence of significant causal relevance between synthetic labels and sample features, which is detrimental to generalization.
One commonly cited example involves the application of the CutMix~\cite{yun2019CutMix} technique, wherein a composite image depicting a dog in the sky is generated by merging segments from a dog situated in a grassy environment and a bird flying in the sky.
In this scene, a robust model should recognize the object ``dog'' by disentangling from the background. 
We highlight that ConCutMix may mitigate potential spurious correlations, because semantically consistent labels share a similar idea to focus on the foreground object. On the other hand, if both dog and bird exist in the sky after CutMix~\cite{yun2019CutMix}, semantically consistent labels can provide training signals for both classes due to the TopK scheme in \eqref{eq:topk}, sharing a similar behavior with area-based labels.
Fig.~\ref{fig:cam} presents empirical evidences that support the viewpoint that semantically consistent labels play a beneficial role in directing the model's attention towards the discriminative foreground. 
In contrast, the adoption of area-based labels like CutMix~\cite{yun2019CutMix} or CMO~\cite{park2022majority} results in model distraction caused by spurious correlations arising from the background.
In light of this observation, we believe that ConCutMix holds promise in mitigating spurious correlation.
}

\begin{figure*}[ht]
\resizebox{\linewidth}{!}{
    \begin{minipage}[t]{0.25\linewidth}
        \centering
        \includegraphics[width=\textwidth]{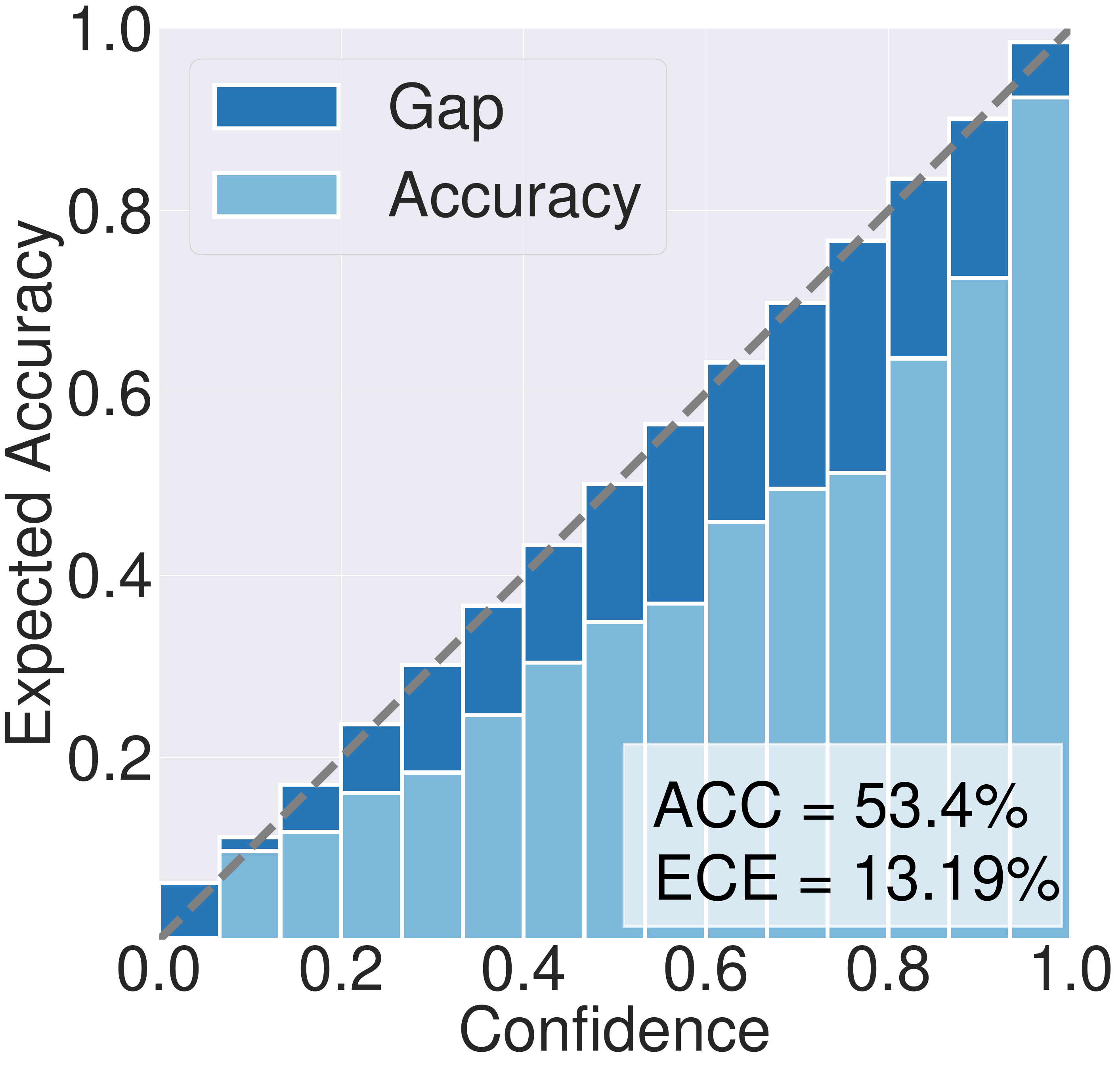}
        \centerline{ \quad BCL}
    \end{minipage}%
    \begin{minipage}[t]{0.25\linewidth}
        \centering
        \includegraphics[width=\textwidth]{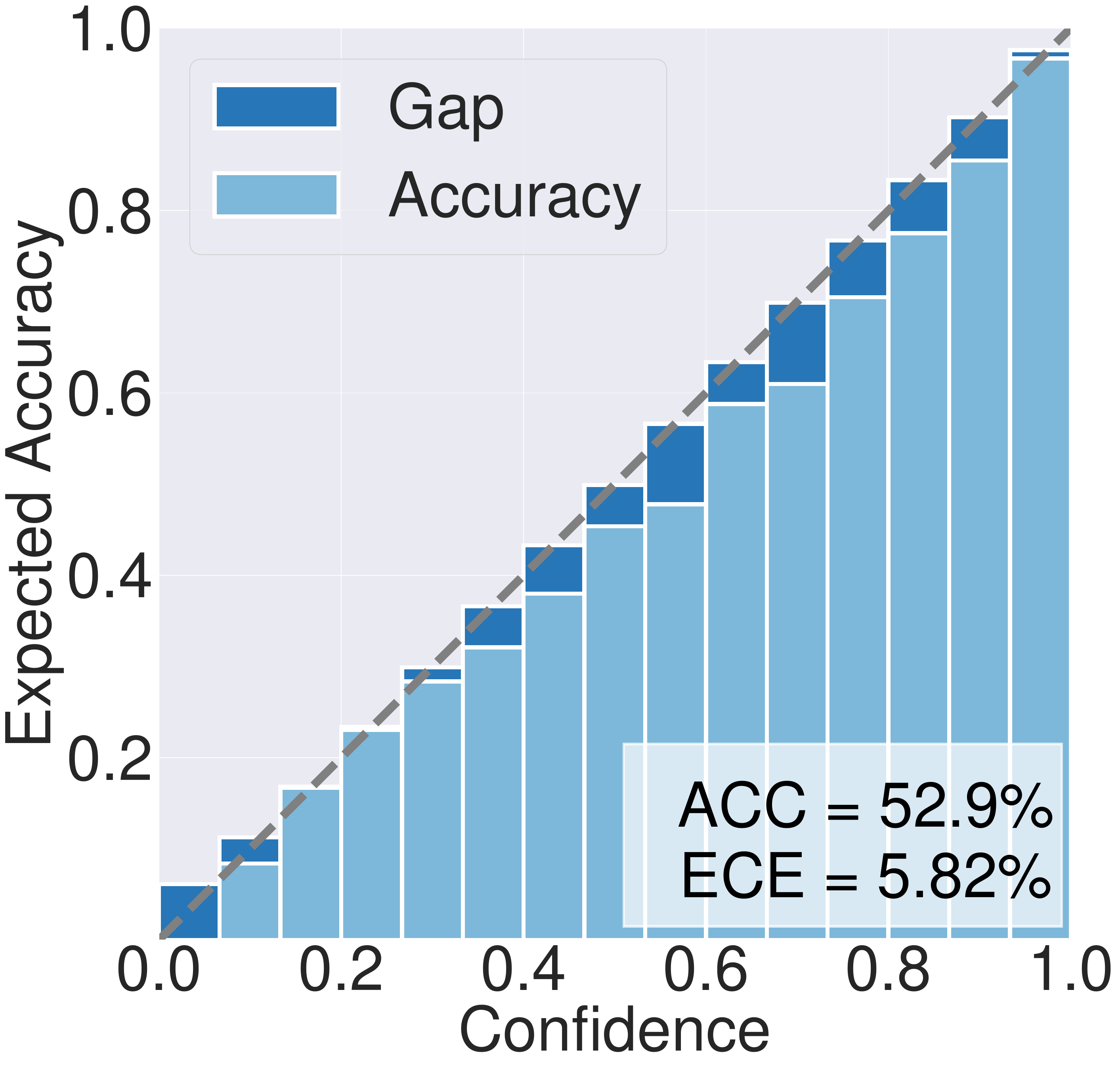}
        \centerline{ \quad CutMix}
    \end{minipage}
    \begin{minipage}[t]{0.25\linewidth}
        \centering
        \includegraphics[width=\textwidth]{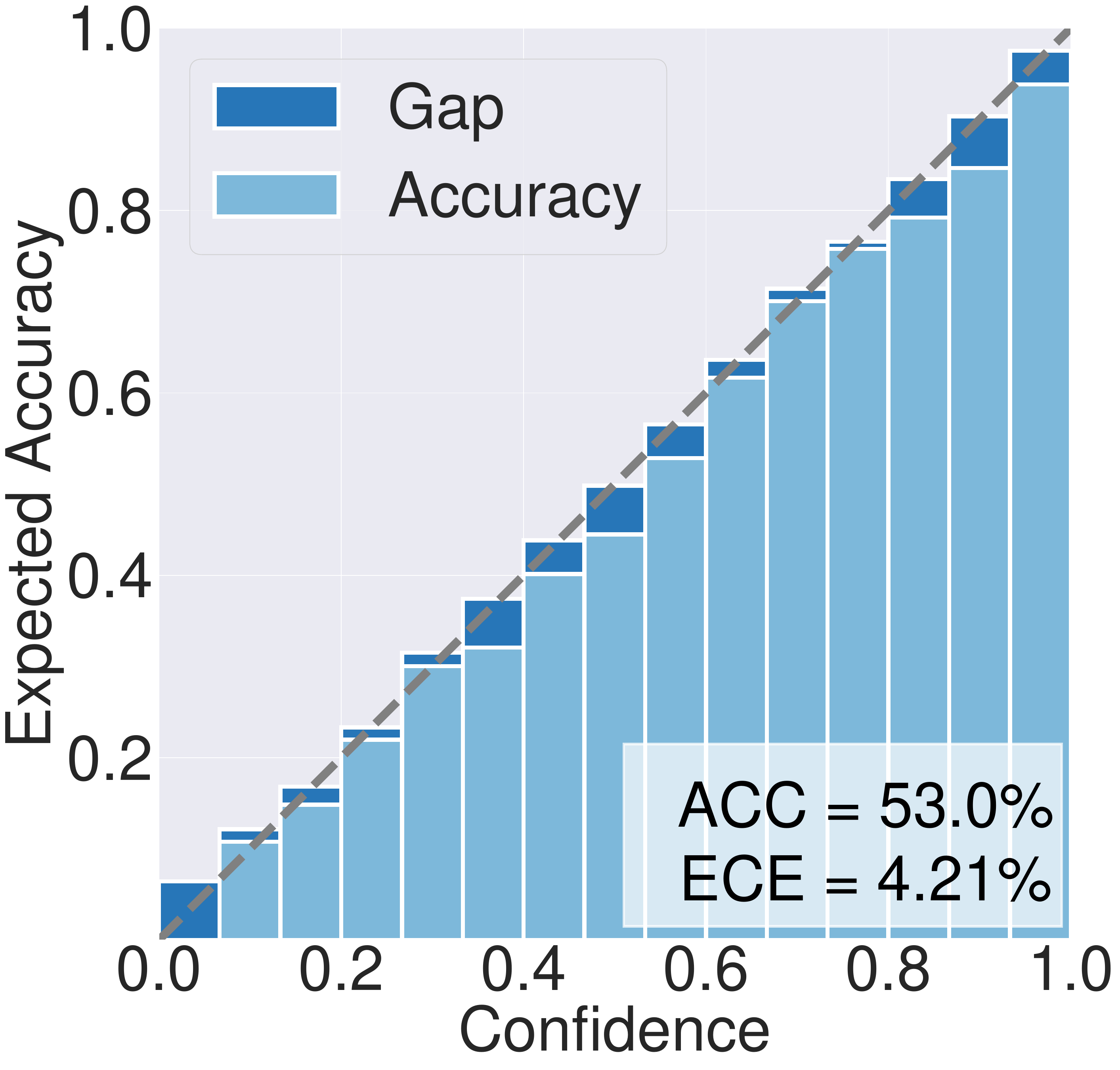}
        \centerline{ \quad CMO}
    \end{minipage}
        \begin{minipage}[t]{0.25\linewidth}
        \centering
        \includegraphics[width=\textwidth]{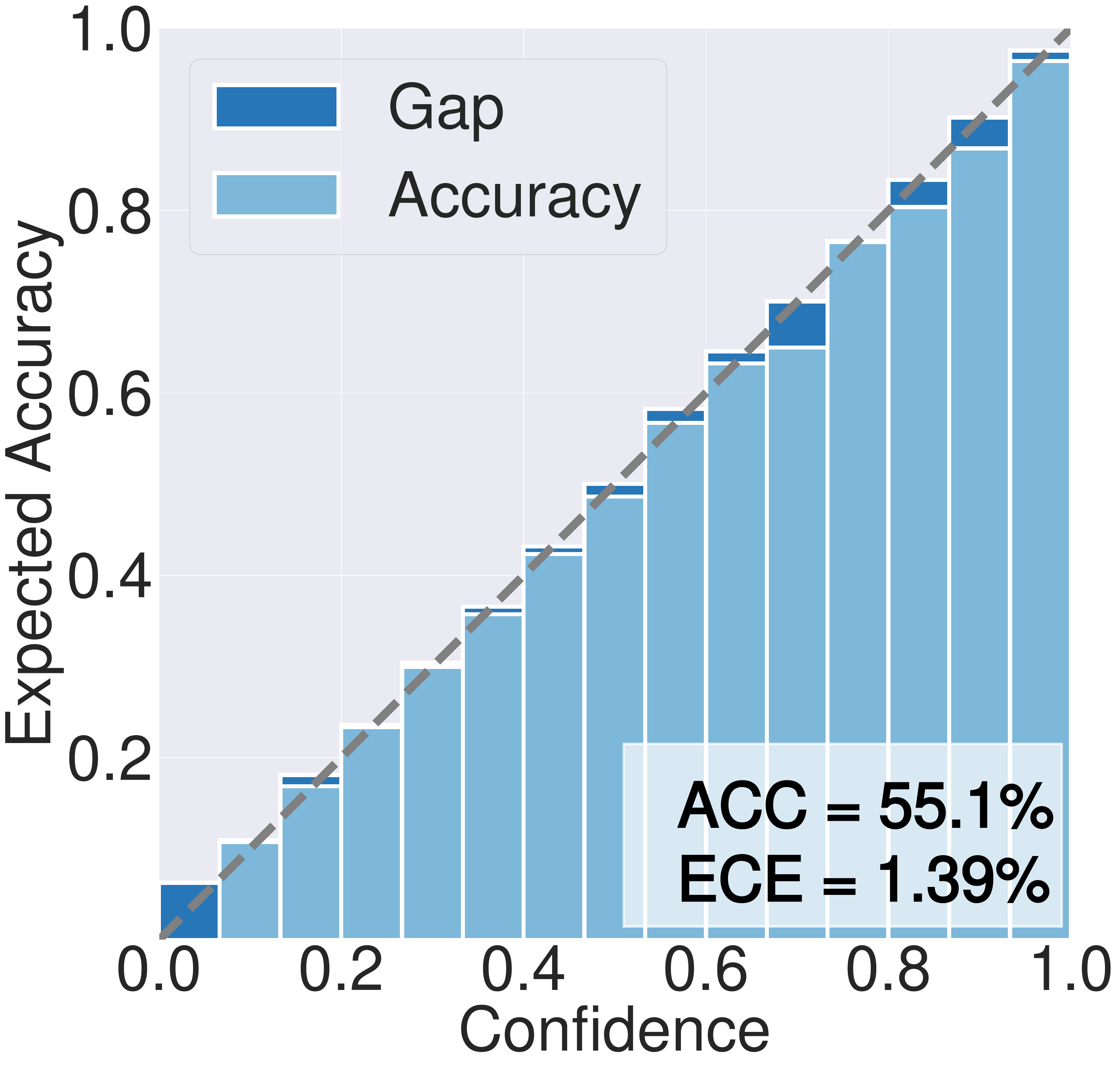}
        \centerline{ \quad ConCutMix (Ours)}
    \end{minipage}
    }

    \resizebox{\linewidth}{!}{
    \begin{minipage}[t]{0.25\linewidth}
        \centering
        \includegraphics[width=\textwidth]{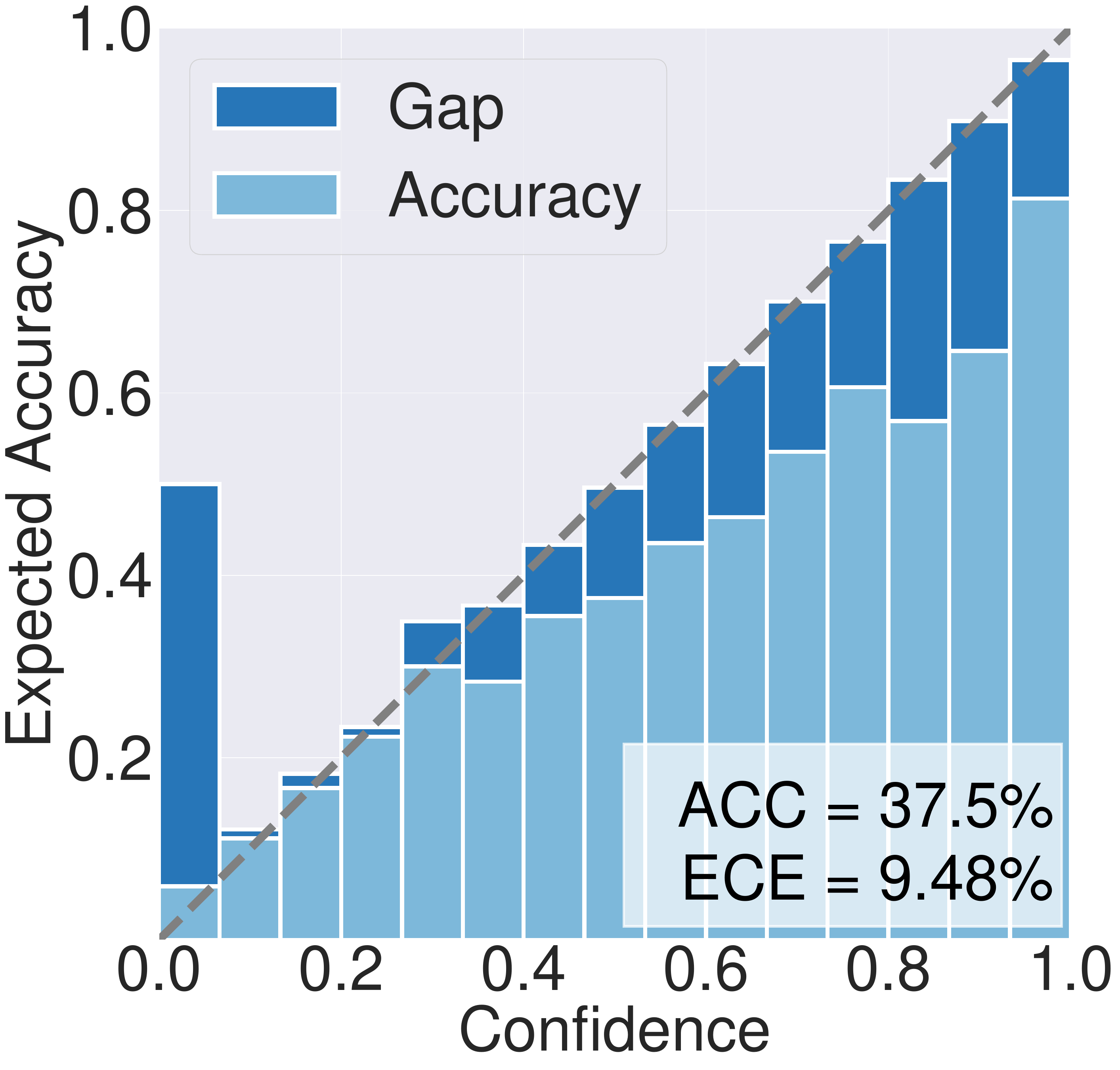}
        \centerline{ \quad BCL {on Tail Classes}}
    \end{minipage}%
    \begin{minipage}[t]{0.25\linewidth}
        \centering
        \includegraphics[width=\textwidth]{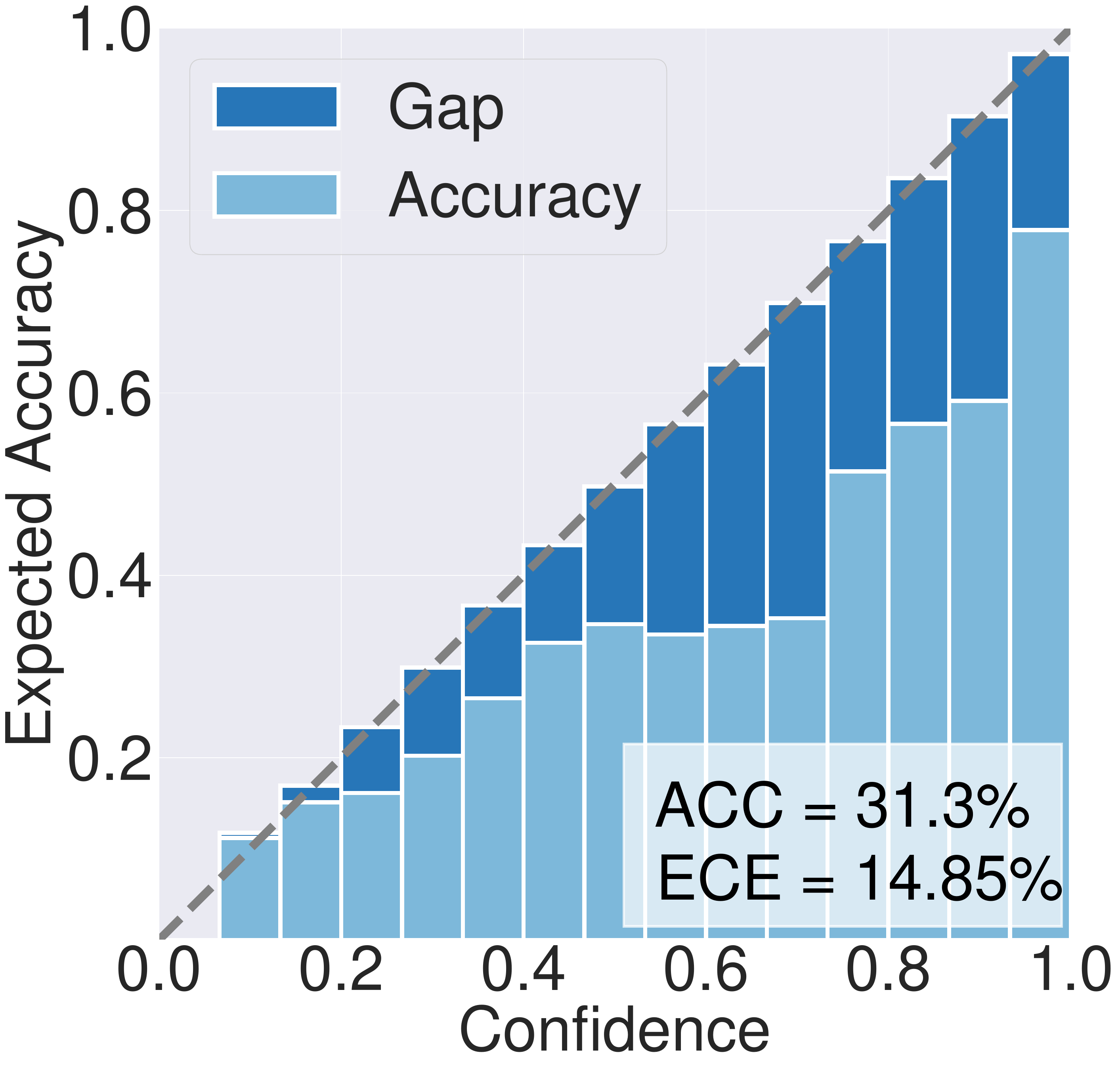}
        \centerline{ \quad CutMix {on Tail Classes}}
    \end{minipage}
    \begin{minipage}[t]{0.25\linewidth}
        \centering
        \includegraphics[width=\textwidth]{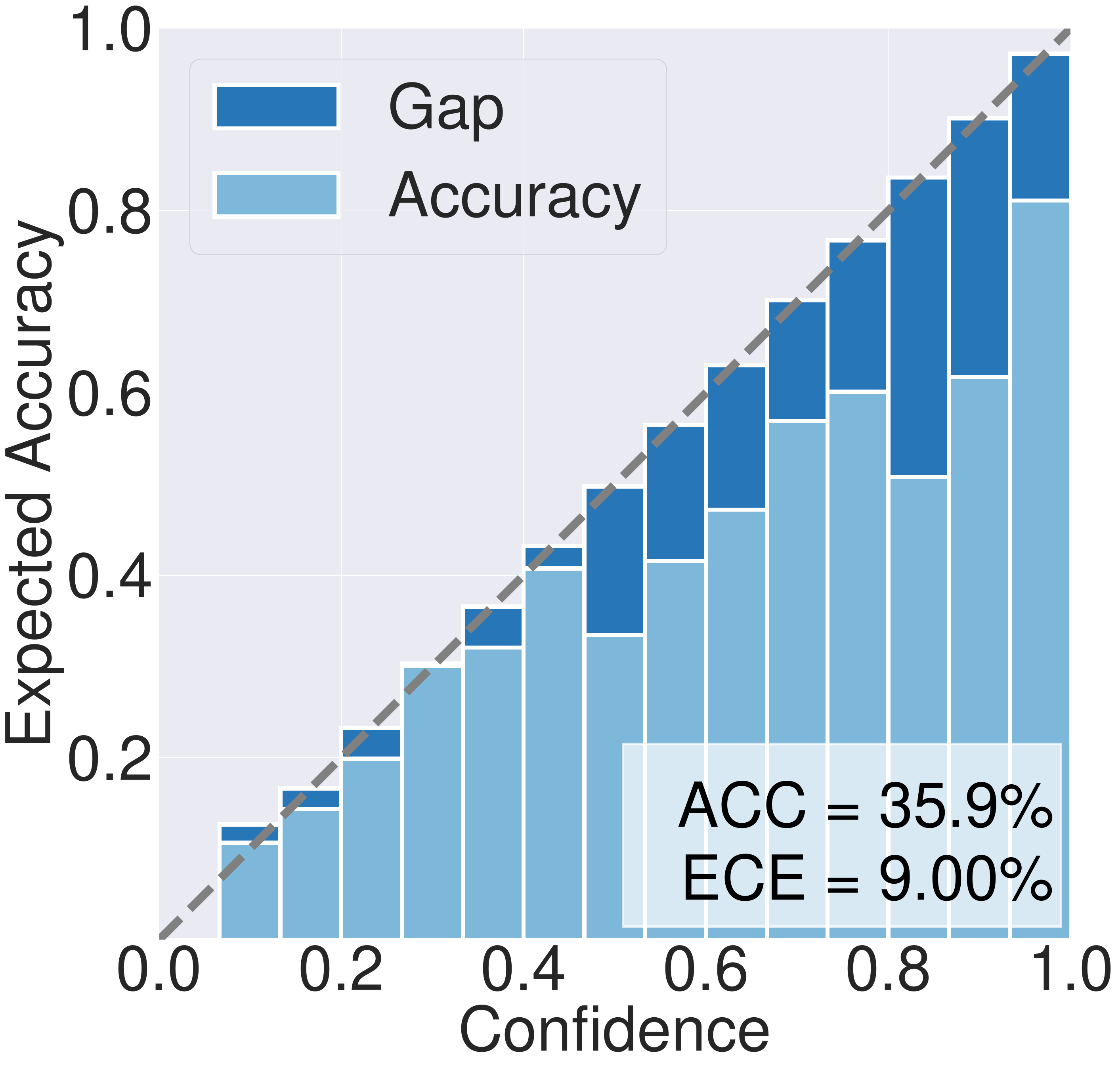}
        \centerline{ \quad CMO {on Tail Classes}}
    \end{minipage}
        \begin{minipage}[t]{0.25\linewidth}
        \centering
        \includegraphics[width=\textwidth]{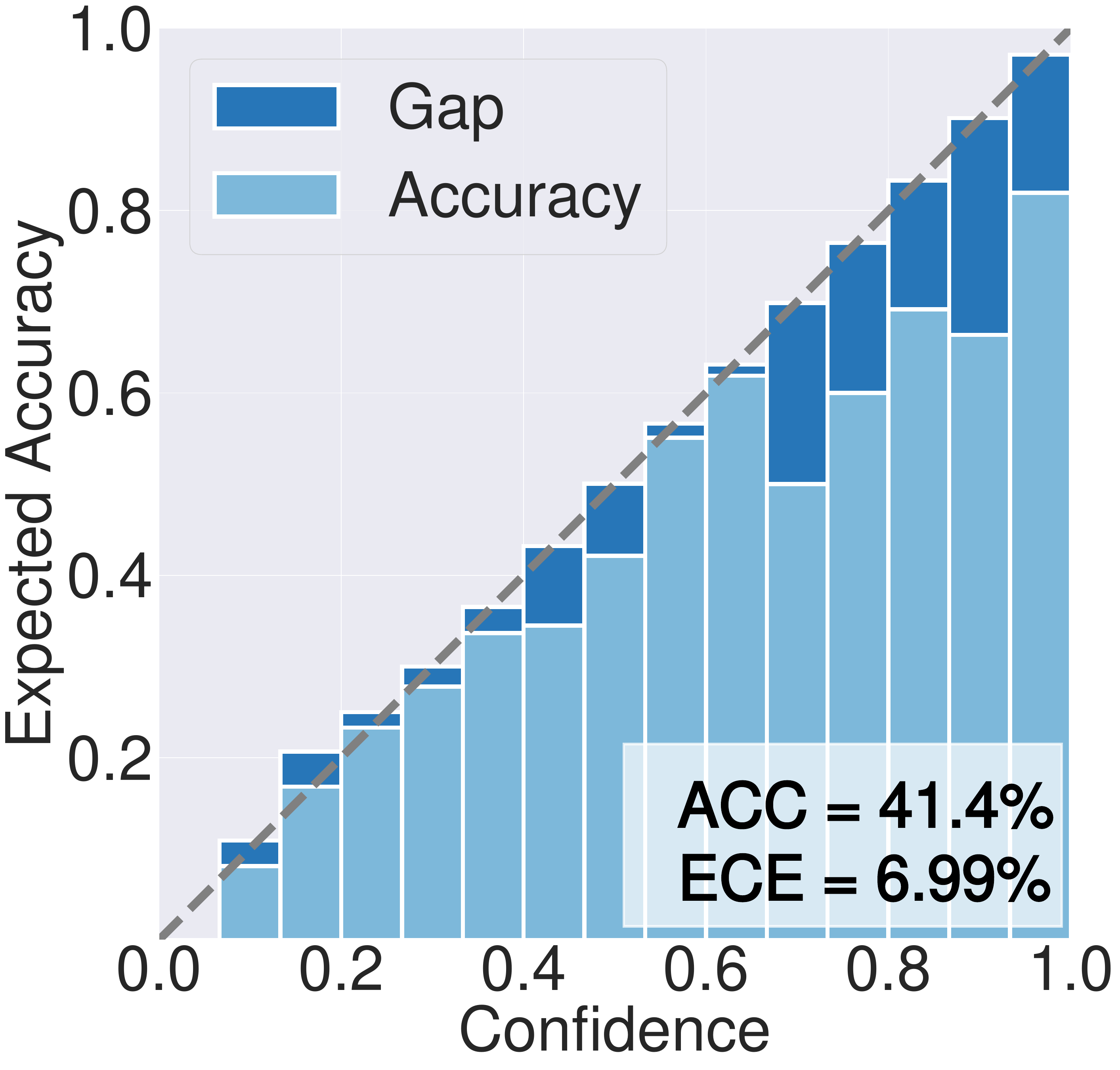}
        \centerline{ \quad ConCutMix {on Tail Classes}}
    \end{minipage}
    }    
    
    \caption{Comparisons of the reliability diagrams of all classes and tail classes among different methods on CIFAR-100-LT at 100 imbalance factor for 400 epochs. Both CutMix~\cite{yun2019CutMix} and CMO~\cite{park2022majority} enables better calibration than BCL~\cite{zhu2022balanced} on all groups.
    However, it shows the damage to the accuracy caused by the noisy training signals.
    As for few-shot group, CutMix~\cite{yun2019CutMix} does not improve the performance of BCL \cite{zhu2022balanced} on the tail classes and show overconfidence for the prediction. CMO~\cite{park2022majority} can reduce overconfidence to some extent, but damage the accuracy due to overfitting the head classes. By Contrast, our ConCutMix always improves the accuracy and reduces the overconfidence of the model, achieving better model calibration.
    }
    \label{fig:reliability_diagrams}
\end{figure*}

\noindent\textbf{Reduced Confusion between Head and Tail Classes.}
To clearly show where the models are getting confused on long-tailed data, we visualize the confusion matrices of predictions of different methods in Fig. \ref{fig:confusion matrix}, and highlight the observed regions of confusion by the \red{red} boxes.
BCL \cite{zhu2022balanced} 
performs well on most classes, but still suffers from confusing head classes samples as tail classes. 
CutMix~\cite{yun2019CutMix} significantly improves the accuracy of head-classes, but 
its area-based labels exacerbate the overfitting to the head-classes, confusing many tail classes samples as head classes.
CMO~\cite{park2022majority} improves the CutMix~\cite{yun2019CutMix} by oversampling the tail class and partially reduces the confusion of samples between head classes and tail classes, but it significantly sacrifices the accuracy of head classes, thereby limiting overall performance.
In contrast, ConCutMix successfully discriminates between the head classes and tail classes, while maintaining the accuracy of the head classes.
Consequently, ConCutMix boosts the overall performance, elevating it from 84.31\% to 86.07\%.

\noindent\textbf{Improved Consistency between Model Predictions and Empirical Probabilities.}
{
In real-world application, aside from accuracy, another significant attribute of a model is its reliability, also known as model calibration, which indicates the consistency between model predictions and empirical probabilities.
\cite{guo2017calibration} proposed that modern models should provide corresponding confidences for predictions to measure model calibration.
\cite{zhong2021improving} demonstrated that the models trained on long-tailed datasets often exhibit one type of miscalibration due to the imbalance: overconfidence, where the confidences are higher than the empirical probabilities.
Interestingly, Thulasidasan \textit{et al.}~\cite{thulasidasan2019mixup} found that mixup enables better model calibration which inspires the follow-up CutMix~\cite{yun2019CutMix} and CMO~\cite{park2022majority}.
The calibration can be measured by Expected Calibration Error (ECE)~\cite{abdar2021review}. Given $N$ predictions that are first grouped into $B$ interval bins of equal size, ECE is formally defined as:}
\begin{equation}
    \mathrm{ECE}=\sum_{b=1}^{B} \frac{\left|\mathcal{S}_{b}\right|}{N}\left|\operatorname{acc}\left(\mathcal{S}_{b}\right)-\operatorname{conf}\left(\mathcal{S}_{b}\right)\right| \times 100 \%,
\end{equation}
where $S_b$ is the set of samples whose prediction scores fall into Bin-$b$. $\operatorname{acc}(\cdot)$ and $\operatorname{conf}(\cdot)$ are the functions used to calculate accuracy and predicted confidence, respectively. A lower ECE means better model calibration.

\begin{table*}[ht]
\caption{
{Comparisons of Top-1 accuracy of different methods implemented with ResNeXt-50 on ImageNet-LT for 90 epochs and 180 epochs. Data augmentation methods always benefit the tail classes, but sometimes sacrifice the accuracy of the head classes. In contrast, ConCutMix brings significant improvements to nearly all groups of classes, yielding better overall performance.
The values in parentheses show the accuracy improvements of ConCutMix over BCL \cite{zhu2022balanced}. 
}}\label{90-table}
\begin{center}

\resizebox{0.65\linewidth}{!}{
\begin{tabular}{l|cccc}
\hline
 Method  & Many & Medium & Few &All \\
\hline
    \multicolumn{5}{c}{90 epochs} \\
    \hline

    Focal Loss \cite{lin2017focal} & 64.3  & 37.1  & 8.2   & 43.7 \\
    $\tau$-norm \cite{kang2019decoupling} & 59.1  & 46.9  & 30.7  & 49.4 \\
    BALMS \cite{ren2020balanced} & 62.2  & 48.8  & 29.8  & 51.4 \\
    LWS \cite{kang2019decoupling}  & 60.2  & 47.2  & 30.3  & 49.9 \\
    Casual Model \cite{tang2020long}& 62.7  & 48.8  & 31.6  & 51.8 \\

     LADE \cite{hong2021disentangling}  & 62.3  & 49.3  & 31.2  & 51.9 \\  
     Disalign  \cite{zhang2021distribution} & 62.7  & 52.1  & 31.4  & 53.4 \\

    BCL \cite{zhu2022balanced}   & 67.2  & 53.9  & 36.5  & 56.7 \\
    ~ + Mixup \cite{zhang2017mixup} &67.6&53.9&39.9& 57.2\\
    ~ + CutMix \cite{yun2019CutMix}  &68.5&54.9&38.8&57.9 \\
    ~ + CMO \cite{park2022majority}  &66.9 &54.7 & 39.3&57.1\\
    ~ + ConCutMix (Ours)  & \textbf{70.7} (\textbf{+3.5}) &  \textbf{56.6} (\textbf{+2.7})&    \textbf{39.8} (\textbf{+3.3})& \textbf{59.7} (\textbf{+3.0}) \\
    \hline
    \multicolumn{5}{c}{180 epochs} \\
    \hline

    LADE \cite{hong2021disentangling} & 65.1  & 48.9  & 33.4  & 53.0 \\
    BALMS \cite{ren2020balanced} & 65.8  & 53.2  & 34.1  & 55.4 \\ 
    PaCo \cite{cui2021parametric} & 64.4  & 55.7  & 33.7  & 56.0 \\

    BCL \cite{zhu2022balanced}   & 66.9  & 54.7  & 38.6  & 57.2 \\
    ~+ Mixup \cite{zhang2017mixup} & 67.6& 54.6& 38.3&57.3 \\
    ~+ CutMix \cite{yun2019CutMix} & 67.7 &56.2& 39.0 &58.3\\
    ~+ CMO \cite{park2022majority} & 65.3& 55.9& 39.2&57.2 \\
    ~+ ConCutMix (Ours)&  \textbf{71.5} (\textbf{+4.6})  &  \textbf{57.2} (\textbf{+2.5})  &  \textbf{40.0} (\textbf{+1.4})   &  \textbf{60.3} (\textbf{+3.1})   \\
    \hline
\end{tabular}}
\end{center}
\end{table*}

{
In order to investigate the impact of ConCutMix on model calibration, we first employ CutMix~\cite{yun2019CutMix}, CMO~\cite{park2022majority} and our ConCutMix on BCL~\cite{zhu2022balanced}. Subsequently, we visualize the predicted confidence and the gap to the expected accuracy for all classes on CIFAR-100-LT in Fig.~\ref{fig:reliability_diagrams} (top row), which can be seen as the reliability diagrams of models.
The results reveal that the BCL \cite{zhu2022balanced} exhibits significant overconfidence and miscalibration.
Such a miscalibration can be easily reduced by all considered data augmentation methods, i.e., achieving lower ECE. 
It is worth noting that the improvements in model calibration achieved by CutMix~\cite{yun2019CutMix} and CMO~\cite{park2022majority} both come at the expense of sacrificing overall accuracy.
In contrast, our ConCutMix not only assists in reducing the ECE from 13.19\% to 1.39\% in the BCL~\cite{zhu2022balanced}, thereby achieving the best model calibration among the considered augmentation techniques, but also leads to a noteworthy accuracy improvement of 1.7\%.
Similarly, as depicted in Fig.~\ref{fig:reliability_diagrams} (bottom row), the reliability diagrams for the tail classes can be drawn in the same way.
We find that CutMix~\cite{yun2019CutMix} and CMO~\cite{park2022majority} result in poorer model calibration or lower accuracy for the tail classes, 
which can be attributed to their inability to mitigate overfitting to the head classes when training for 400 epochs on CIFAR-100-LT.
These results highlight the effectiveness of our ConCutMix in both alleviating miscalibration and boosting recognition performance to the classes with scarce samples.}

\noindent\textbf{Better Trade-off between Tail and Head Classes.}
{
Following~\cite{zhu2022balanced}, we report the results of different groups where there are significant differences in the number of samples, i.e., Many-shot (\textgreater 100 images), Medium-shot (20$\sim$100 images), and Few-shot (\textless20 images). 
As shown in Table~\ref{90-table}, when training model on the large-scale dataset, ImageNet-LT for 90 epochs, all data augmentation methods benefit the accuracy of the baseline on Few-shot group.
However, in comparison to CutMix~\cite{yun2019CutMix}, we observed that CMO~\cite{park2022majority} achieves higher accuracy on Few-shot group but leads to significant decline in accuracy on Many-shot group.
This observation can be attributed to the idea of CMO~\cite{park2022majority} that emphasising tail classes to improve CutMix~\cite{yun2019CutMix}.
Consequently, a trade-off between head and tail classes occurs, involving the improvement of tail class accuracy at the expense of head class accuracy.
This trade-off resembles a "seesaw" effect and is often observed in long-tailed methods that aim to rebalance class distributions~\cite{deeplongtail}, resulting in a limited overall performance for CMO~\cite{park2022majority}.
Notably, when the number of training epochs is increased to 180, the sacrifice in accuracy of head classes becomes more pronounced for CMO~\cite{park2022majority}, resulting in a 1.1\% decrease in overall performance compared to CutMix~\cite{yun2019CutMix}.
In contrast, the proposed ConCutMix improves upon CutMix~\cite{yun2019CutMix} by the idea of rectifying noisy area-based labels by semantically consistent labels. 
As a result, ConCutMix enhances the quality and quantity of synthetic samples of tail classes while preserving the performance of head classes.
Experimental results indicate that ConCutMix consistently achieves better trade-off between tail and head classes and leads to significant improvements across all class groups, outperforming other methods in terms of overall performances for both 90 and 180 epochs.
}

\begin{figure*}[t]
    \centering
\centerline{\includegraphics[width=\linewidth]{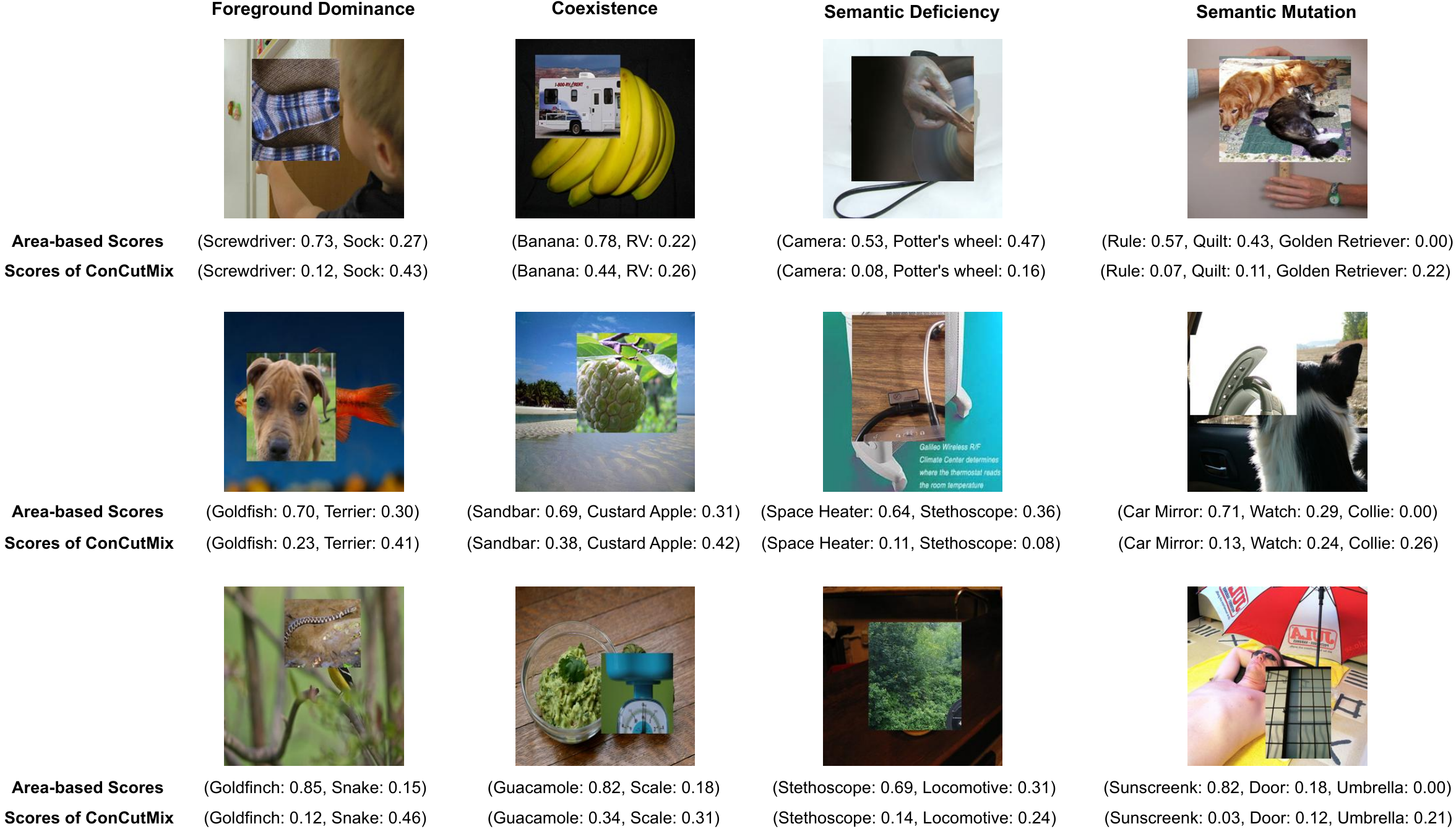}}    \caption{
{More semantic samples and scores are compared between ConCutMix and area-based scores. ConCutMix  demonstrates good visual effects across various scenarios, including Foreground Dominance (Left 1), Coexistence (Left 2), Semantic Deficiency (Right 2), Semantic Mutation (Right 1).}}
    \label{fig:examples}
\end{figure*}

{
\noindent \textbf{Providing better scores across diverse mixing scenarios.}
We include more examples that visually contrast the scores of ConCutMix with area-based scores in Fig. \ref{fig:examples}, showcasing ConCutMix provides better scores across diverse mixing scenarios. We show examples contrasting ConCutMix and area-based scores, highlighting ConCutMix's superior semantic assessment of synthetic samples.
1) Foreground Dominance: The semantic content of the foreground covers the background, resulting in an image with only the foreground’s semantic information. Area-based scoring inaccurately assigns a higher score to the larger background area. ConCutMix correctly assigns a higher score to the foreground and a lower score to the occluded background. 2) Coexistence: Both the foreground and background contribute their semantic information to the CutMix image. ConCutMix effectively assigns high scores to both classes, similar to area-based scoring, but with an emphasis on each class's semantic contribution. 3) Semantic Deficiency: Both the foreground and background lack clear semantic information, leading to an ambiguous image. Area-based scores are guided by area coverage, which can be misleading. ConCutMix lowers the scores for both, aligning better with the perception of minimal informative content. 4) Semantic Mutation: The foreground obscures the background, potentially creating a new class. Area-based scoring might prioritize the background due to its larger visible area. ConCutMix, however, prioritizes the more dominant foreground, reflecting the actual visual input. }

\section{Conclusion}
In this paper, we investigate the problem of long-tail recognition from the perspective of data augmentation. 
We discover that the area-based labels generated by CutMix ignore the semantic information of synthetic samples, thus producing noisy training signals. 
To address this issue, we propose a Contrastive CutMix augmentation (ConCutMix) method which use the semantically consistent labels to rectify the area-based labels.
{To address this issue, we propose a Contrastive CutMix augmentation (ConCutMix) method which uses the semantically consistent labels to rectify the area-based labels.}
Specifically, we first explicitly compute the similarities between synthetic samples with class centers in a semantic space learned by contrastive learning. Then, we consider TopK-similar classes to construct semantically consistent labels to capture the semantic information of novel classes that are not used in CutMix.
Finally, we incorporate semantically consistent labels into training under the control of the confidence function.
Extensive experiments show that the proposed ConCutMix significantly improves the performance on all the considered benchmarks.

\section{Acknowledgments}
This work is supported by the National Natural Science Foundation of China (Grant No. 62376099 \& 62072186) and Natural Science Foundation of Guangdong Province (Grant No. 2024A1515010989).

\bibliographystyle{IEEEtran}
\bibliography{ref.bib}

\end{document}